\documentclass[preprint,12pt,authoryear]{elsarticle}

\usepackage{amssymb}

\usepackage{color}
\usepackage{url}
\usepackage{algorithm}
\usepackage{algpseudocode}
\usepackage{hyperref}
\usepackage{subfig}
\usepackage[section]{placeins}
\usepackage{todonotes}
\usepackage{amsmath}
\usepackage{glossaries}
\usepackage{duckuments}
\usepackage[capitalize,noabbrev]{cleveref}

\journal{Neural Networks}

\glsdisablehyper
\newacronym{APL}{APL}{%
  the Johns Hopkins University Applied Physics Laboratory%
}

\newacronym{ACC}{ACC}{average task accuracy}

\newacronym{AUC}{AUC}{area under the curve}

\newacronym{A2C}{A2C}{advantage actor critic}

\newacronym{BT}{BT}{backward transfer}

\newacronym{DFCNN}{DF-CNN}{deconvolutional factorized convolutional neural network}

\newacronym{EB}{EB}{evaluation block}
\newacronym{EX}{EX}{evaluation experience}

\newacronym{FT}{FT}{forward transfer}

\newacronym{JHU}{JHU}{%
  Johns Hopkins University%
}

\newacronym{LL}{LL}{Lifelong Learning}

\newacronym{LPGFTW}{LPG-FTW}{lifelong policy gradients for faster training without forgetting}

\newacronym{L2M}{L2M}{DARPA Lifelong Learning Machines}

\newacronym{LX}{LX}{learning experience}

\newacronym{LB}{LB}{learning block}

\newacronym{MDP}{MDP}{Markov decision process}

\newacronym{METAKFO}{Meta-KFO}{meta Kronecker factorization optimization}

\newacronym{ML}{ML}{machine learning}

\newacronym{PM}{PM}{Performance Maintenance}

\newacronym{RL}{RL}{reinforcement learning}
\newacronym{RP}{RP}{performance relative to a Single Task Expert}

\newacronym{SE}{SE}{sample efficiency}

\newacronym{STE}{STE}{single-task expert}

\newacronym{STELLAR}{STELLAR}{Super Turing Evolving Lifelong Learning ARchitecture}

\newacronym{TE}{T\&E}{test and evaluation}

\newacronym{TDY}{TDY}{Teledyne}

\newacronym{UML}{UML}{Uncertainty-Modulated Learning}

\begin{document}

\begin{frontmatter}

%% Title, authors and addresses

\title{A Domain-Agnostic Approach for Characterization of Lifelong Learning Systems}

\author[org:apl]{ {\bf Megan M. Baker}} % APL
\author[org:apl]{ {\bf Alexander New}} % APL
\author[org:tdy]{ {\bf Mario Aguilar-Simon}} % Teledyne
\author[org:uta]{ {\bf Ziad Al-Halah}} % UT Austin
\author[org:usc]{ {\bf Sébastien M. R. Arnold}} % USC
\author[org:lu]{ {\bf Ese Ben-Iwhiwhu}} % Loughborough
\author[org:tdy]{ {\bf Andrew P. Brna}} % Teledyne
\author[org:umich]{ {\bf Ethan Brooks}} % Michigan
\author[org:tdy]{ {\bf Ryan C. Brown}} % Teledyne
\author[org:sri]{ {\bf Zachary Daniels}} % SRI
\author[org:utsa]{ {\bf Anurag Daram}} %UTSA
\author[org:umass]{ {\bf Fabien Delattre}} % UMass
\author[org:snl]{ {\bf Ryan Dellana}} %Sandia
\author[org:upenn]{ {\bf Eric Eaton}} % UPenn
\author[org:brown]{ {\bf Haotian Fu}} % Brown
\author[org:uta]{ {\bf Kristen Grauman}} % UT Austin
\author[org:sri]{ {\bf Jesse Hostetler}} % SRI
\author[org:usc]{ {\bf Shariq Iqbal}} % USC
\author[org:upenn]{ {\bf Cassandra Kent}} % UPenn
\author[org:hrl]{ {\bf Nicholas Ketz}} % HRL
\author[org:vb]{ {\bf Soheil Kolouri}} % Vanderbilt
\author[org:brown]{ {\bf George Konidaris}} % Brown
\author[org:utsa]{ {\bf Dhireesha Kudithipudi}} %UTSA
\author[org:umass]{ {\bf Erik Learned-Miller}} % UMass
\author[org:upenn]{ {\bf Seungwon Lee}} % UPenn
\author[org:brown]{ {\bf Michael L. Littman}} % Brown
\author[org:anl]{ {\bf Sandeep Madireddy}} %Argonne
\author[org:upenn]{ {\bf Jorge A. Mendez}} % UPenn
\author[org:apl]{ {\bf Eric Q. Nguyen}} % APL
\author[org:apl]{ {\bf Christine Piatko}} % APL
\author[org:hrl]{ {\bf Praveen K. Pilly}} % HRL
\author[org:sri]{ {\bf Aswin Raghavan}} % SRI
\author[org:sri]{ {\bf Abrar Rahman}} % SRI
\author[org:uta]{ {\bf Santhosh Kumar Ramakrishnan}} % UT Austin
\author[org:hrl]{ {\bf Neale Ratzlaff}} % HRL
\author[org:lu]{ {\bf Andrea Soltoggio}} % Loughborough
\author[org:uta]{ {\bf Peter Stone}} % UT Austin
\author[org:sri]{ {\bf Indranil Sur}} % SRI
\author[org:umass]{ {\bf Zhipeng Tang}} % UMass
\author[org:brown]{ {\bf Saket Tiwari}} % Brown
\author[org:upenn]{ {\bf Kyle Vedder}} % UPenn
\author[org:snl]{ {\bf Felix Wang}} %Sandia
\author[org:uta]{ {\bf Zifan Xu}} % UT Austin
\author[org:anl]{ {\bf Angel Yanguas-Gil}} %Argonne
\author[org:uta]{ {\bf Harel Yedidsion}} % UT Austin
\author[org:brown]{ {\bf Shangqun Yu}} % Brown
\author[org:apl]{ {\bf Gautam K. Vallabha}} % APL

% Affiliation Definitions
\affiliation[org:apl]{organization={Johns Hopkins University Applied Physics Laboratory},%Department and Organization
            addressline={11100 Johns Hopkins Rd.}, 
            city={Laurel},
            postcode={20723}, 
            state={MD},
            country={USA}}
\affiliation[org:tdy]{organization={Teledyne Scientific Company - Intelligent Systems Laboratory},%Department and Organization
            addressline={19 T.W. Alexander Drive}, 
            city={RTP},
            postcode={27709}, 
            state={NC},
            country={USA}}
\affiliation[org:uta]{organization={Department of Computer Science, University of Texas at Austin},%Department and Organization
            addressline={}, 
            city={Austin},
            postcode={}, 
            state={TX},
            country={USA}}
\affiliation[org:usc]{organization={Department of Computer Science, University of Southern California},%Department and Organization
            addressline={}, 
            city={Los Angeles},
            postcode={}, 
            state={CA},
            country={USA}}
\affiliation[org:lu]{organization={Department of Computer Science, Loughborough University},%Department and Organization
            addressline={}, 
            city={Loughborough},
            postcode={}, 
            state={England},
            country={UK}}  
\affiliation[org:umich]{organization={Department of Electrical Engineering and Computer Science, University of Michigan},%Department and Organization
            addressline={}, 
            city={Ann Arbor},
            postcode={}, 
            state={MI},
            country={USA}}
\affiliation[org:sri]{organization={SRI International},%Department and Organization
            addressline={201 Washington Rd}, 
            city={Princeton},
            postcode={}, 
            state={NJ},
            country={USA}}
\affiliation[org:utsa]{organization={University of Texas at San Antonio},%Department and Organization
            addressline={}, 
            city={San Antonio},
            postcode={}, 
            state={TX},
            country={USA}}
\affiliation[org:umass]{organization={Department of Computer Science, University of Massachusetts Amherst},%Department and Organization
            addressline={}, 
            city={Amherst},
            postcode={}, 
            state={MA},
            country={USA}}
\affiliation[org:snl]{organization={Sandia National Laboratories},%Department and Organization
            addressline={}, 
            city={Albuquerque},
            postcode={}, 
            state={NM},
            country={USA}}
\affiliation[org:upenn]{organization={Department of Computer and Information Science, University of Pennsylvania},%Department and Organization
            addressline={}, 
            city={Philadelphia},
            postcode={}, 
            state={PA},
            country={USA}}
\affiliation[org:brown]{organization={Department of Computer Science, Brown University},%Department and Organization
            addressline={}, 
            city={Providence},
            postcode={}, 
            state={RI},
            country={USA}}  
\affiliation[org:hrl]{organization={Information and Systems Sciences Laboratory, HRL Laboratories},%Department and Organization
            addressline={3011 Malibu Canyon Road}, 
            city={Malibu},
            postcode={90265}, 
            state={CA},
            country={USA}}
\affiliation[org:vb]{organization={Department of Computer Science, Vanderbilt University},%Department and Organization
            addressline={}, 
            city={Nashville},
            postcode={}, 
            state={TN},
            country={USA}}
\affiliation[org:anl]{organization={Argonne National Laboratory},%Department and Organization
            addressline={9700 S Cass Ave}, 
            city={Lemont},
            postcode={}, 
            state={IL},
            country={USA}}

\begin{abstract}
Despite the advancement of machine learning techniques in recent years, state-of-the-art systems lack robustness to ``real world'' events, where the input distributions and tasks encountered by the deployed systems will not be limited to the original training context, and systems will instead need to adapt to novel distributions and tasks while deployed. This critical gap may be addressed through the development of ``Lifelong Learning'' systems that are capable of 1) \textit{Continuous Learning}, 2) \textit{Transfer and Adaptation}, and 3) \textit{Scalability}. Unfortunately, efforts to improve these capabilities are typically treated as distinct areas of research that are assessed independently, without regard to the impact of each separate capability on other aspects of the system. We instead propose a holistic approach, using a suite of metrics and an evaluation framework to assess Lifelong Learning in a principled way that is agnostic to specific domains or system techniques. Through five case studies, we show that this suite of metrics can inform the development of varied and complex Lifelong Learning systems. We highlight how the proposed suite of metrics quantifies performance trade-offs present during Lifelong Learning system development - both the widely discussed Stability-Plasticity dilemma and the newly proposed relationship between Sample Efficient and Robust Learning. Further, we make recommendations for the formulation and use of metrics to guide the continuing development of Lifelong Learning systems and assess their progress in the future.
\end{abstract}

\end{frontmatter}
\footnotetext[1]{Distribution Statement ``A'' (Approved for Public Release, Distribution Unlimited)}
\section{Introduction}\label{sec:introduction}

While \gls{ML} has made dramatic advances in the past decade, deployment and use of data-driven \gls{ML}-based systems in the real world faces a crucial challenge: the input distributions and tasks encountered by the deployed system will not be limited to the original training context, and systems will need to accommodate novel distributions and tasks while deployed. We define the challenge of \gls{LL} as enabling a system to learn and retain knowledge of multiple tasks over its operational lifetime. Addressing this challenge requires new approaches to both algorithm development and assessment.
The \gls{L2M} program was initiated in 2018 to stimulate fundamental algorithmic advances in \gls{LL} and to assess these \gls{LL} capabilities in complex environments.  The program focused on both \gls{RL} and classification systems in diverse domains, such as CARLA~\citep{Dosovitskiy17CARLA} (3D simulator for autonomous driving), StarCraft~\citep{StarCraft} (real-time strategy game), AI Habitat~\citep{habitat19iccv} (photorealistic 3D simulator for indoor environments), AirSim~\citep{Airsim2017} (3D drone simulator), and L2Explorer~\citep{Johnson2022L2Explorer} (open-world exploration). The diversity of domains was motivated primarily by the research consideration of exploring \gls{LL} in a broad array of contexts, and it resulted in each research team developing \gls{LL} systems for their respective domains.  

Throughout this work, we use the term “\gls{LL} system” rather than “\gls{LL} algorithm”, as the developed systems were composed of many different interacting components (e.g. regularization, experience replay, task change detection, etc.). The capability to do \gls{LL} is a property of the overall system rather than any one component, and multiple metrics are needed to characterize \gls{LL} systems.

The evaluation of these \gls{LL} systems faced two key questions: (1) what metrics are most suitable for assessing \gls{LL}, and (2) how can one apply these \gls{LL} Metrics in a consistent way to different \gls{LL} systems, each operating in a different domain? In particular, a primary purpose of this evaluation was to measure progress over the course of the program and to assess the strengths and weaknesses of different systems in an environment-agnostic manner, thereby providing deeper insight into \gls{LL}.

The rest of this paper is organized as follows: In~\cref{sec:background}, we give an overview on \gls{LL} systems, as well as different approaches for evaluating them. In~\cref{sec:metric_evaluation}, we introduce the core components of our approach for evaluating \gls{LL}--conditions of \gls{LL}, evaluation scenarios, and evaluation protocols. In~\cref{sec:metrics}, we define the metrics we use to evaluate \gls{LL} systems. In~\cref{sec:case_studies}, we describe a set of case studies that demonstrate the application of these metrics to varied domains. In~\cref{sec:discussion}, we conclude with insights from these case studies and give recommendations for assessing and advancing \gls{LL} systems.  Throughout this work, we introduce and use a number of terms which are defined in ~\ref{sec:terminology}.

\section{Background}\label{sec:background}
The area of machine \gls{LL} has recently seen a large amount of attention in the research community~\citep{Silver2013Lifelong,Chen2018Book,Parisi2019LL,Hadsell2020review,delange2021continual}, especially through its connections to other subfields such as multi-task~\citep{Caruana1997multitask,Zhang2021multitask}, transfer~\citep{Zhuang2019transfer}, incremental batch~\citep{Kemker2018forgetting}, and online~\citep{Hoi2018online} learning; as well as domain adaptation~\citep{Csurka2017domainadaptation} and generalization~\citep{Zhou2022domaingeneralization}. The distinguishing characteristic of \gls{LL} is that a deployed system encounters a sequence of tasks over its lifetime, with no prior knowledge of the number, structure, duration, or re-occurrence probability of those tasks. The two key challenges are to retain expertise on previously learned tasks, thereby avoiding catastrophic forgetting \citep{McCloskey1989CF,Ratcliff1990forgetting,French1992catastrophic,French1999forgetting,mcclelland1995there,Goodfellow2013forgetting}, and to transfer acquired expertise to facilitate learning of new tasks~\citep{Pratt1991transfer,Sharkey1993transfer}. Ultimately, an ideal \gls{LL} system leverages relationships among tasks to improve performance across all tasks it encounters, even if the input distributions of those tasks change over a lifetime. Earlier work considered the challenges of developing algorithms to avoid forgetting and enhance transfer~\citep{Pratt1992transfer,Ring1997child}.

\begin{figure}
    \centering
    \includegraphics[width=0.8\linewidth, scale=0.75]{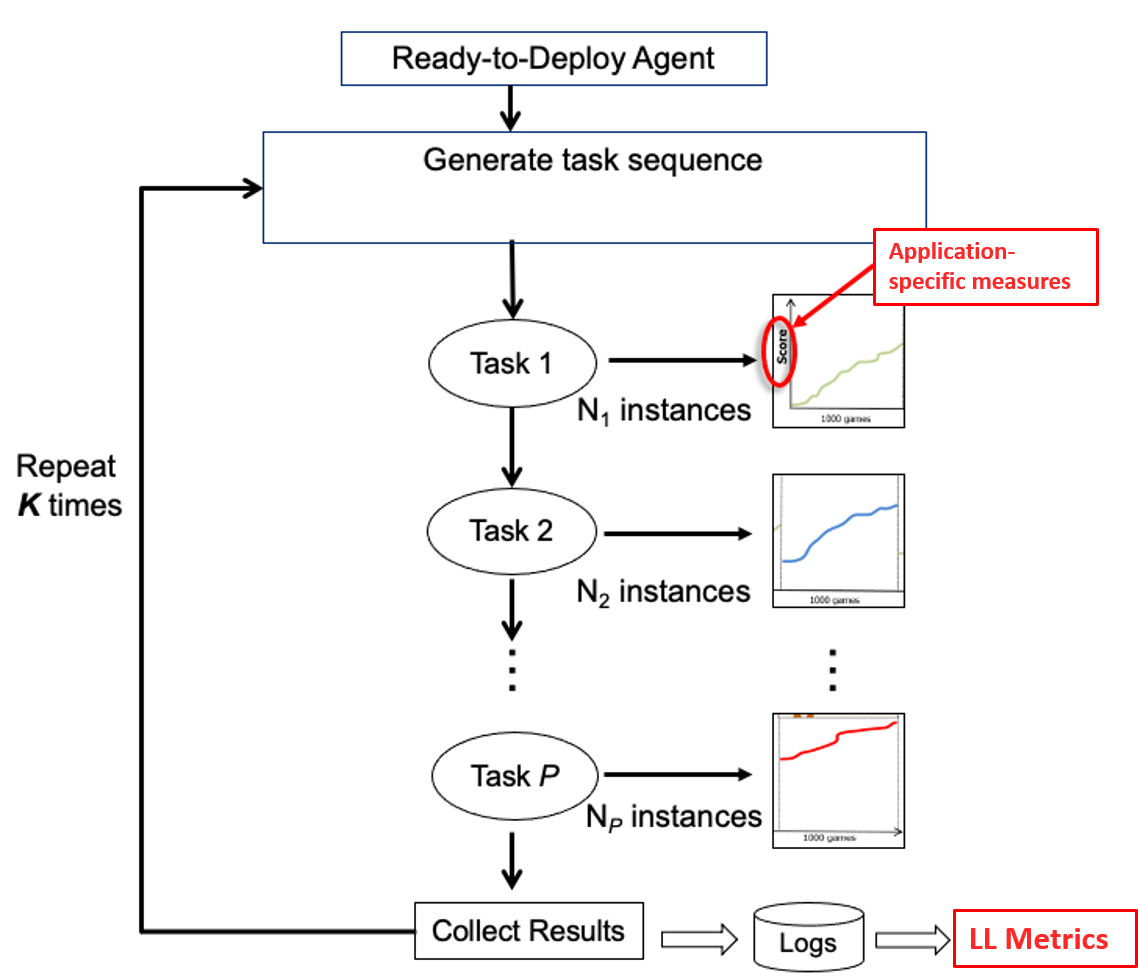}
    \caption{Depiction of a \gls{LL} Scenario generated to evaluate a given \gls{LL} system according to the approach outlined in this work. The \gls{LL} Scenario shown here and described in ~\cref{sec:metric_evaluation} is an environment-agnostic template used to define the number and sequence of tasks, how they are sequenced in a given “lifetime” (or run) of the \gls{LL} system, and how many repeats should be performed to generate statistically reliable results. These sequences of tasks generate application-specific measures (see ~\cref{sec:application-specific}) that feed into the calculation of \gls{LL} Metrics, shown in red and defined in ~\cref{sec:metrics}. The \gls{LL} Metrics track performance within a system's lifetime, and are best interpreted in the context of the corresponding \gls{LL} Scenario. Examples of this analysis and the impact the suite of \gls{LL} Metrics provide can be found in ~\cref{sec:case_studies}, followed by practical considerations and insights for assessing and advancing \gls{LL} systems in ~\cref{sec:discussion}. See~\ref{sec:terminology} for further definitions of the terms used here.}
    \label{fig:scenario-design}
\end{figure}

As different methods and algorithms for \gls{LL} have been developed, various approaches have been taken for evaluating these systems. A key distinction has been made between evaluation scenarios and metrics: evaluation scenarios (as shown in~\cref{fig:scenario-design}) set up the structure of the lifetime of the \gls{LL} system--what tasks occur, how they are presented, and how often--whereas metrics assess how well the system performed over that lifetime. We recommend~\citet{mundt2022clevacompass} as a concurrently-developed work focusing on the challenges of categorizing different \gls{LL} algorithms and evaluations in terms of transparency, replicability, and contextualization. When constructing a set of metrics, it is important to decide what they should be assessing. \cite{Zhu2020TransferSurvey} frame metrics for \gls{LL} as assessing either generalization (how prior knowledge facilitates initial learning on a new task) or mastery (how prior knowledge facilitates eventual performance on a new task). The suite of metrics defined in this paper extends these concepts by defining conditions of \gls{LL} in Section~\ref{sec:conditions}.

\subsection{Evaluation Scenarios for Different Learning Paradigms}\label{sec:paradigms}
The difficulty of quantitatively evaluating \gls{LL} systems has led to a variety of approaches, both specific to the learning type and more general.
Quantitatively assessing the performance of classification \gls{LL} systems is often more straightforward than assessing \gls{RL} systems because there are straightforward ways of generating tasks from a dataset (e.g., by splitting sets of classes into tasks, or by inducing domain shifts). However, while evaluating the \gls{LL} capability of a classification system is still challenging, the evaluation scenarios used to do so tend to be specific to the classification context, such as incremental class learning, e.g.,~\cite{Hsu2018categorization}.
Despite this, there are broader insights that are applicable for \gls{RL} as well, as noted by  ~\cite{farquhar2019robust}. In particular, ~\cite{Hayes2018NewParadigms} identify different methods of setting up the sequence of observations that constitute each lifetime of the system: sampling from different tasks in an \emph{i.i.d.} fashion, grouping them by task or by class labels within a task, or (most challenging) sampling and grouping them in a non-\emph{i.i.d.} fashion.

Evaluation of lifelong \gls{RL} faces additional challenges: (1) \gls{RL} can be highly variable within and across training runs, and across rollouts of a fixed policy \citep{Chan2020Reliability}, (2) rewards across different tasks may have different scales or extrema, or may be unbounded, and (3) it is nontrivial to design tasks with well-characterized relationships (see, e.g.,~\cite{Carroll2005tasksimilarity}). Nonetheless, work on \gls{RL} generalization and transfer offers valuable insight for \gls{LL}. ~\cite{Kirk2021Generalization} propose a useful formalism of a ``contextual \gls{MDP}'' where for each episode encountered by the system, the state of the \gls{MDP} encodes an unseen “context” (e.g., random seeds and parameters used to specify the task). During training and test, the system encounters episodes sampled from training and test context sets respectively, with generalization assessed using zero-shot forward transfer and a “generalization gap” metric (difference in expected rewards between train and test). One of their key recommendations is to specify tasks using a combination of procedural content generation (which varies based on parameters inherent to the environment) and explicitly specified parameters. In CORA, ~\cite{Powers2021CORA} present a different approach for \gls{RL} performance assessment. They handcrafted benchmark tasks for four different environments (Atari~\citep{bellemare13arcade}, ProcGen~\citep{cobbe2020procgen}, MiniHack~\citep{samvelyan2021minihack} and AI2-Thor~\citep{Kolve2017ai2thor}), and proposed a standard evaluation protocol ($N$ tasks presented sequentially, cycled $M$ times).

\subsection{Metrics for Different Learning Paradigms}\label{sec:paradigms}

Metrics commonly used to assess the performance of classification \gls{LL} systems include \gls{ACC}, \gls{FT} and \gls{BT} (also denoted FWT and BWT, respectively), as well as model size, storage and computational efficiency \citep{DiazRodriguez2018DontForget, lopez2017gradient}.  Other metrics specifically developed for classification \gls{LL} include Cumulative Gain, which tracks \gls{ACC} after each task exposure during the course of the system's lifetime \citep{BenavidesPrado2020CumulativeGain}, $\Omega_{all}$, an extension of \gls{ACC} that compares the accuracy to an offline learner \citep{Hayes2018NewParadigms}, and Performance Drop \citep{balaji2020effectiveness}, which uses the baseline of a multi-task model trained jointly on all tasks.

Metrics used for assessing lifelong \gls{RL} include those introduced by ~\cite{Powers2021CORA} for use in CORA: Forgetting (change in performance on a task before and after learning a new task) and zero-shot \gls{FT} (change in performance after learning a new task, relative to a random agent). They also present baseline algorithms demonstrating the value of the metrics and tasks. ~\cite{Zhu2020TransferSurvey} also propose metrics for two-task transfer learning, comparing performance with and without prior task exposure: initial performance, asymptotic performance, accumulated reward (measured by an \gls{AUC} calculation), and time to a threshold performance. They also propose a Transfer Ratio (asymptotic performance measured as a ratio), and performance sensitivity (variance in performance with different hyperparameter settings). 

In summary, there is currently no clear guidance for defining tasks or scenarios to exercise LL, other than the guidance of having multiple tasks with some kind of structured similarity and presenting tasks to the system without specifying the order beforehand. There are also no universally accepted metrics for \gls{LL}, though \gls{FT} is often used for both classification and \gls{RL}, and average (or cumulative) change in performance is used in \gls{RL}. Overall, there is no agreed-upon standard for how to assess \gls{LL} systems across different environments in a uniform manner.

\subsection{DARPA L2M Program Context}\label{sec:context}

The \gls{L2M} program was initiated to stimulate fundamental advances in lifelong \gls{ML} systems. Of particular interest were systems operating in complex and challenging environments and potentially applicable to a broad array of domains (including autonomous driving, embodied search, and real-time strategy). To this end, research conducted under the program coalesced into five different domains.

\begin{table}[ht]
\centering
\begin{tabular}{|c|c|c|c|}
\hline
\textbf{\begin{tabular}{@{}c@{}}System Group \\Designation\end{tabular}} & \textbf{Environment} & \textbf{Domain} \\
\hline
SG-UPenn (\ref{sec:sg-upenn}) &\begin{tabular}{@{}c@{}}AI Habitat \\~\citep{habitat19iccv}\end{tabular} &  \begin{tabular}{@{}c@{}}Robotics \\ embodied search\end{tabular}\\
\hline
SG-Teledyne (\ref{sec:sg-teledyne}) & \begin{tabular}{@{}c@{}}AirSim \\\citep{Airsim2017}\end{tabular} &  \begin{tabular}{@{}c@{}}Autonomous navigation \\ (drones)\end{tabular} \\
\hline
SG-HRL (\ref{sec:sg-hrl}) & \begin{tabular}{@{}c@{}}CARLA \\\citep{Dosovitskiy17CARLA}\end{tabular}&  \begin{tabular}{@{}c@{}}Autonomous navigation \\ (cars, motorcycles)\end{tabular} \\
\hline
SG-Argonne (\ref{sec:sg-argonne}) & \begin{tabular}{@{}c@{}}L2Explorer\\~\citep{Johnson2022L2Explorer}\end{tabular}  &  \begin{tabular}{@{}c@{}}Open-world \\ exploration\end{tabular}  \\
\hline
SG-SRI (\ref{sec:sg-sri}) & \begin{tabular}{@{}c@{}}StarCraft 2 \\\citep{StarCraft}\end{tabular}&  \begin{tabular}{@{}c@{}}Game play / \\ real-time strategy\end{tabular} \\
\hline
\end{tabular}
\caption{Five \gls{LL} systems were developed during the \gls{L2M} Program, and the teams were led by the organizations listed. The corresponding environment and domain are shown. The variation in the domains represented in the \gls{L2M} Program necessitated the development of domain- and environment-agnostic metrics, as well as \gls{LL} threshold values at which a system is said to be exhibiting Lifelong Learning. These domains can include classification and/or Reinforcement Learning components.}
\label{tab:ll-systems}
\end{table}
\cref{tab:ll-systems} provides information on the five \gls{LL} systems that were developed as part of the program, along with their associated environments/domains.  In this work, we focused on the evaluation of systems within these five environments, but the concepts and methods are broadly applicable and could work well in conjunction with a library like Avalanche ~\citep{Lomonaco2021Avalanche}. We treated each \gls{LL} system as a black box, intentionally omitting details of the constituent components. Each system was developed by a different research team and their algorithmic advances are described in publications contained in \cref{sec:case_studies}.

\subsection{Evaluation of LL systems}
How exactly to assess such a wide variety of \gls{LL} systems operating in diverse environments was a major challenge addressed during the course of the \gls{L2M} Program. We emphasize that the goal was not to identify the “best” \gls{LL} system, as each environment required different learning strategies. Instead, the goal was to provide deeper insight into the strengths and weaknesses of \gls{LL} systems in an environment-agnostic manner.
The \gls{L2M} Program \gls{TE} team and research teams collaboratively identified and defined the following key components of an LL evaluation:

\begin{enumerate}
    \item The \textbf{Conditions of \gls{LL}} the system needed to demonstrate, which are defined in ~\cref{sec:conditions}. These conditions specify diverse criteria identifying different components of the overall phenomena of \gls{LL}.
    \item The \textbf{Evaluation Scenarios} that exercise the \gls{LL} system for the purpose of computing metrics. This is an environment-agnostic template that defined the number of tasks and constraints on their relationships, as well as how they are sequenced in a given “lifetime” (or run) of the \gls{LL} system. An example is demonstrated in~\cref{fig:scenario-design} and details are provided in ~\cref{sec:ll_scenarios}.
    \item The overall \textbf{Evaluation Protocol} specifies how multiple lifetimes are set up, and consists of the Evaluation Scenarios as well as details (e.g. number of lifetimes) for obtaining statistically reliable metrics. Evaluation Protocols are discussed in Section~\ref{sec:evaluation_protocol}.
    \item The set of \textbf{\gls{LL} Metrics} (described in ~\cref{sec:metrics}) that assess the conditions of \gls{LL}. We discovered early on that a single metric would not be sufficient to cover all the conditions, and multiple metrics would be needed to characterize the \gls{LL} systems.
\end{enumerate}

\section{Evaluation Approach}\label{sec:metric_evaluation}

We consider three key aspects of evaluating \gls{LL} systems--the conditions of \gls{LL} (\cref{sec:conditions}), scenarios that systems encounter (\cref{sec:ll_scenarios}), and the overall protocols that specify an evaluation (\cref{sec:evaluation_protocol}).

\subsection{Conditions of Lifelong Learning}\label{sec:conditions}
We assert that an \gls{LL} system must satisfy three necessary and sufficient conditions: 

\begin{enumerate}
    \item \textbf{Continuous Learning:} The \gls{LL} system learns a nonstationary stream of tasks (both novel and recurring), continually consolidating new information to improve performance while coping with irrelevance and noise.
    \item \textbf{Transfer and Adaptation:} As learning progresses, the \gls{LL} system performs better on average on the next task it experiences, for both novel and known tasks (forward and backward transfer), maintaining performance during rapid changes in the ongoing task (adaptation).
    \item \textbf{Scalability:} The \gls{LL} system continues learning for an arbitrarily long lifetime using limited resources (e.g., memory, time) in a scalable way.
\end{enumerate}

These three conditions of \gls{LL} have been used to drive the development of \gls{LL} Metrics. They are similar to the notion of `generalization' and `mastery' introduced by \cite{Zhu2020TransferSurvey}, and two of our metrics can measure these concepts. The jumpstart formulation of \gls{FT} (a Transfer and Adaptation metric) can be considered a measure of `generalization,' and \gls{RP} - a Scalability metric - can be considered a measure of `mastery.'
It is important to point out that these conditions are partially independent; indeed, it is possible for a system to demonstrate \gls{LL} in one condition but not in another. Because of this, it is all the more critical to use multiple measures to assess \gls{LL} systems.  The relationship between the Metrics, the Conditions of LL, and Scenario requirements associated with assessing them are discussed further in~\cref{sec:metrics}.

It is also worth noting the relationship between the above definition and related terms such as ``Continual Learning'' \citep{Chen_Liu_2018}. There are two aspects here. First, are the learning experiences from different tasks intermixed as an \textit{i.i.d} sequence (online or streaming learning \citep{Hayes_Cahill_Kanan_2018}) or as a non-\textit{i.i.d} sequence with same-task experiences being batched together? Second, do new learning experiences expand the domain of already-learned tasks (incremental class learning), or are they entirely new tasks with new input and output domains (incremental task learning) \citep{vandeVen_Tolias_2018}?

Lifelong Learning, as defined above, is incremental task learning with same-task experiences batched together and with the additional constraint that the system leverage prior knowledge to become a more effective and efficient learner.  The term ``Continual Learning'' has historically been used to loosely refer to either incremental task or class learning. However, over the past few years, it has been used more synonomously with Lifelong Learning. To avoid confusion, we consistently use the term ``Lifelong Learning'' in this paper.

\subsection{Evaluation Scenarios} \label{sec:ll_scenarios}

An Evaluation Scenario describes the patterns and frequency of task or task variant repetitions in sequence, and can facilitate evaluating \gls{LL} systems with respect to specific metrics as well as provide insight into their strengths and weaknesses. Since certain task sequences are required to reasonably explore LL metrics, specifying a particular Scenario is a critical step in characterizing the performance of an \gls{LL} system. 

\begin{figure}[htbp!]
  \centering
  \includegraphics[width=1.0\linewidth]{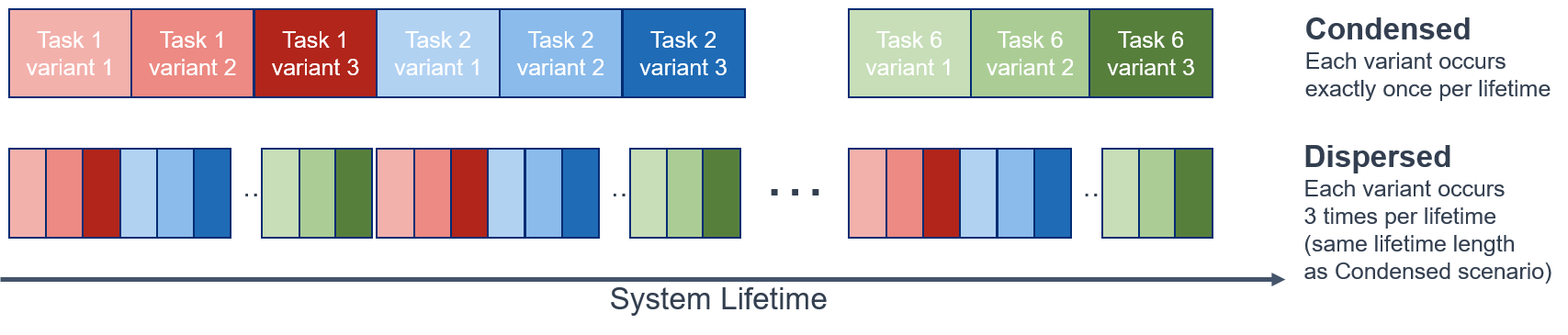}
    \caption{Illustration of Condensed and Dispersed Scenario Types introduced in \cref{sec:ll_scenarios} and used in the case studies of~\cref{sec:case_studies}. The structure of these scenarios was chosen to aid in consistent, thorough evaluation of an \gls{LL} system and to explore how system performances vary based on differences in task ordering and frequency of task switching.}
\label{fig:scenario-types}
\end{figure}

Two of the main scenario types used to accomplish this were Condensed and Dispersed Scenarios. Both scenario types are illustrated in \cref{fig:scenario-types}, with further details in~\ref{sec:supplementary_scenario_info}. Each involved a sequence of multiple tasks and variants. Individual runs had different permutation orders.

In particular, Condensed Scenarios involved concentrating all of the experience per task in one longer block. Dispersed Scenarios involved the same amount of experience per task, but with interleaved tasks in shuffled segments rather than appearing in sequence. These two scenario types were chosen to explore differences in system performance based on task ordering and appearance (since an operationalized system will not have prior knowledge of task sequences), and to ensure enough task repetitions for reasonably evaluating whether a system retained expertise on previously seen tasks. In \cref{sec:case_studies}, we see that some LL systems perform differently in various scenarios. These differences enable us to identify the characteristics, strengths, and weaknesses of an \gls{LL} system.

In developing these scenario structures, we built on existing work in this area. For example,~\cite{vandeVen2019ThreeScenarios} proposed the class-incremental learning scenario, which is similar in structure to our condensed scenario. Concurrently to our work,~\citet{Cossu2021class_incremental} built off this and suggested the class-incremental with repetition scenario, which is similar to our dispersed scenario. Similarly,~\citet{Stojanov2019crib} designs a class-incremental scenario that features parametric variation in its task design. Our framework differs in two key ways from these. First, it is meant to be more general than these scenarios, as it can accommodate \gls{LL} systems that perform classification and/or reinforcement learning. Second, it incorporates task variants into its structure, which can help evaluate \gls{LL} systems on environments with similar sets of tasks. Ultimately, the existence of these other scenarios is beneficial for exploring the combinatorial design space of \gls{LL} scenarios, and benchmarks can be shared and extended. See~\ref{sec:supplementary_scenario_info} for a full example of what an Evaluation Scenario looks like.

\subsection{Evaluation Protocols}\label{sec:evaluation_protocol}
In order to evaluate a particular \gls{LL} system (consisting of a fixed set of hyperparameters, algorithms, and components), we recommend the use of an Evaluation Protocol. An Evaluation Protocol is a complete specification for conducting \gls{LL} Scenarios to ensure reproducibility and obtain statistically reliable \gls{LL} Metrics.

In addition to one or more Evaluation Scenarios, this specification consists of details about pre-deployment training (e.g., pretraining on a fixed dataset like ImageNet), and how multiple lifetimes (runs) should be generated for each scenario. This evaluation approach was used in the \gls{L2M} program to foster experimentation on \gls{LL} Metrics and to help researchers evaluate the performance and progress of their \gls{LL} systems. 

In addition to the Scenario specification, an Evaluation Protocol contains details for obtaining statistically reliable \gls{LL} Metrics. As has been noted in the literature \citep{Agarwal2021deep,Colas2018Power, Colas2019hitchhikers, Henderson2018,Dror2019Dominance}, the training process for deep \gls{RL} systems is noisy and variable, making it challenging to robustly evaluate them.

Our approach to generate statistically reliable \gls{LL} Metrics is based on guidance in \cite{NISTMethods}, and similar to \cite{Colas2018Power}. More details on this approach are provided in ~\ref{sec:reliability}. In contrast to much of the literature, which considers the problem of comparing two or more algorithms, here we focus on the challenge of obtaining reliable estimates of a system's performance (with respect to the metrics defined in~\cref{sec:metrics}). Given such reliable estimates, we are able to determine whether a system is meeting a particular threshold. We further propose the use of \textit{LL thresholds} in \cref{sec:metrics} to determine whether a system is demonstrating LL or not.

\glsresetall

\section{Lifelong Learning Metric Definitions}\label{sec:metrics}

\begin{figure}[H]
    \centering
    \includegraphics[width=\textwidth]{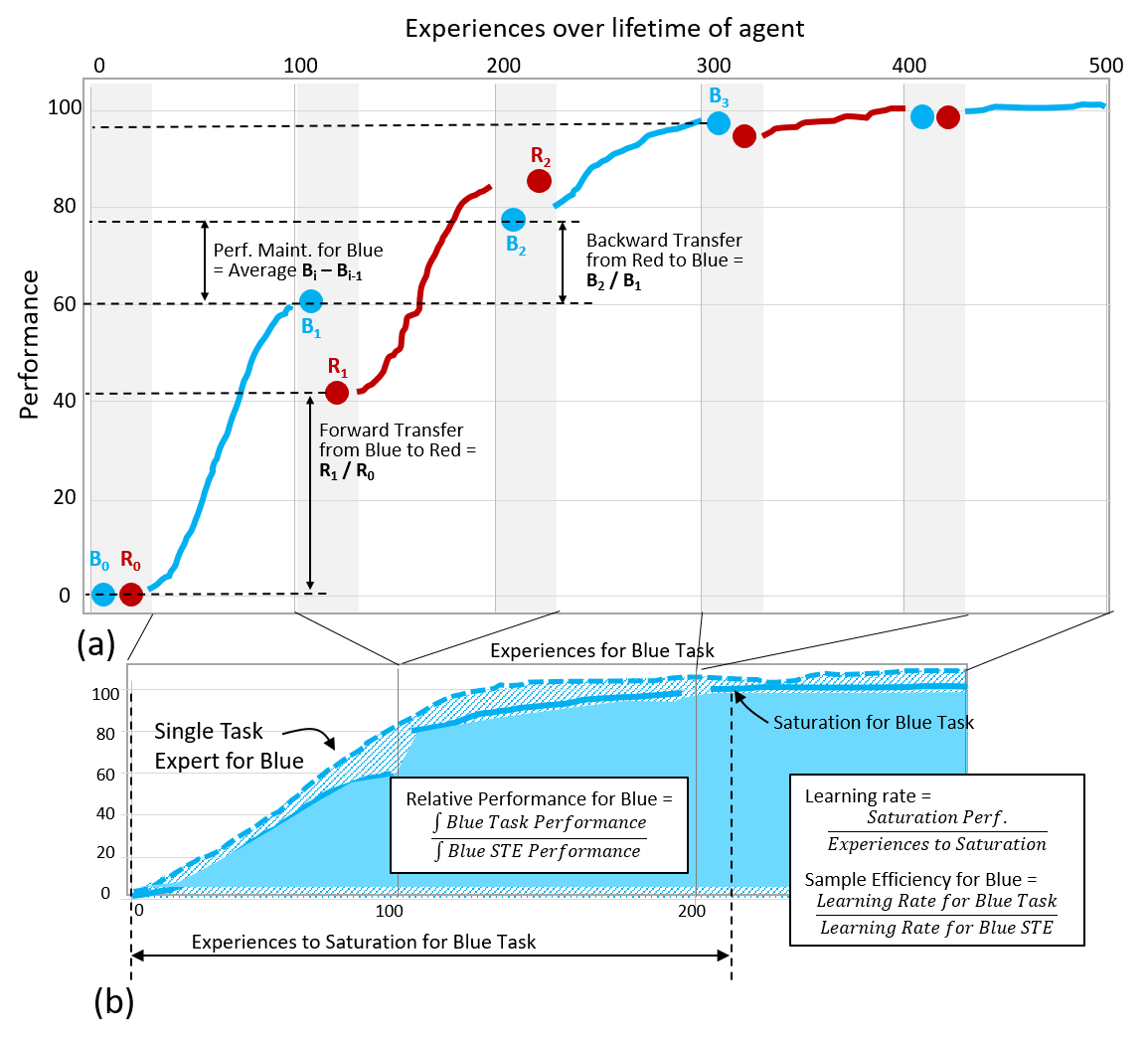}
    \caption{Performance output for an \gls{LL} system in a scenario with two tasks indicated in blue (B) and red (R), annotated to illustrate the computation of the five \gls{LL} Metrics described in this section. \\
    (a) White regions in the graph indicate Learning Blocks, and shaded regions indicate Evaluation Blocks. $B_i$ and $R_i$ refer to performance in the $i$th Evaluation Block for the Blue and Red tasks, respectively. Horizontal dashed lines indicate relevant evaluation performance comparison points referred to in the example formulations of Performance Maintenance, Forward Transfer, and Backward Transfer Metrics. \\
    (b) Single task expert (dashed blue) and \gls{LL} system (solid blue) curves for the scenario shown in Fig. A. Vertical lines indicate the boundaries between each of the three Learning Blocks for the Blue Task stitched from above and overlaid with the Single task expert performance output of the same number of Learning Experiences. Experiences to Saturation and the Saturation Value for the Blue Task are also indicated on the figure to illustrate the example formulations of Sample Efficiency and Relative Performance Metrics.} 
    \label{fig:sawtooth_curve}
\end{figure}

The Lifelong Learning Metrics are scenario, domain, environment and task-agnostic measures that characterize one or more \gls{LL} capabilities across the lifetime of the system.
This suite of \gls{LL} Metrics, summarized in~\cref{table:metrics-summary-table} and visualized in~\cref{fig:sawtooth_curve}, operates on application-specific performance measures (\cref{sec:application-specific}), making the evaluation methodology as separate as possible from the implementation details of a particular system. 

\begin{table}[!ht]
    \centering
    \small
    \begin{tabular}{|c|c|p{0.50\linewidth}|}
    \hline
\textbf{Metric Name} & \textbf{LL Condition} & \textbf{Assesses the LL system's ability to:} \\ \hline
\begin{tabular}{@{}c@{}}Performance\\Maintenance (PM)\end{tabular} & \begin{tabular}{@{}c@{}}Continuous\\Learning\end{tabular} & Avoid catastrophic forgetting despite the introduction of new parameters or  tasks \\ \hline
\begin{tabular}{@{}c@{}}Forward \\Transfer (FT)\end{tabular} & \begin{tabular}{@{}c@{}}Transfer \& \\Adaptation\end{tabular}  & Use expertise in a known task to facilitate learning a new task  \\ \hline
\begin{tabular}{@{}c@{}}Backward \\Transfer (BT)\end{tabular} & \begin{tabular}{@{}c@{}}Transfer \& \\Adaptation\end{tabular} & Use expertise in a new task to improve performance on a known task \\ \hline
\begin{tabular}{@{}c@{}}Relative \\Performance (RP)\end{tabular} & Scalability & Match or exceed the performance of a single-task expert \\ \hline
\begin{tabular}{@{}c@{}}Sample \\Efficiency (SE)\end{tabular} & Scalability & Make better use of learning experiences than an equivalent single-task expert \\ \hline
    \end{tabular}
    \caption{High-level description of the suite of five \gls{LL} Metrics used in this work, described in more detail in Section 4. An in-depth discussion of the specific formulation of the \gls{LL} Metrics can be found in ~\citep{New2022lifelong}.}
    \label{table:metrics-summary-table}
\end{table}

The metrics are meant to work in a complementary manner in order to illustrate and characterize system capability. Thus, there is some overlap in the conditions they measure, as shown in~\cref{table:metrics-summary-table}, as well as in the means employed to do so. This approach ensures that no single metric value is responsible for fully quantifying an \gls{LL} system's performance and instead encourages deeper analysis into specific performance characteristics and the trade-offs between them. 

The relationship between these metrics and the trade-offs illuminated by the case studies in~\cref{sec:case_studies} are explored further in Section~\ref{sec:discussion}. An in-depth discussion of the context of these metrics and their use can be found in ~\cite{New2022lifelong}. Detailed formulations from ~\cite{New2022lifelong} are provided in ~\ref{sec:metrics_formulations}, and a publicly-available Python implementation of the metrics and a logging framework for systems that generate them are available online~\citep{Nguyen2022logger,Nguyen2022metrics}.

\subsection{Application-specific measures}\label{sec:application-specific}

\begin{figure}[htbp!]
  \centering
  \includegraphics[width=0.95\linewidth]{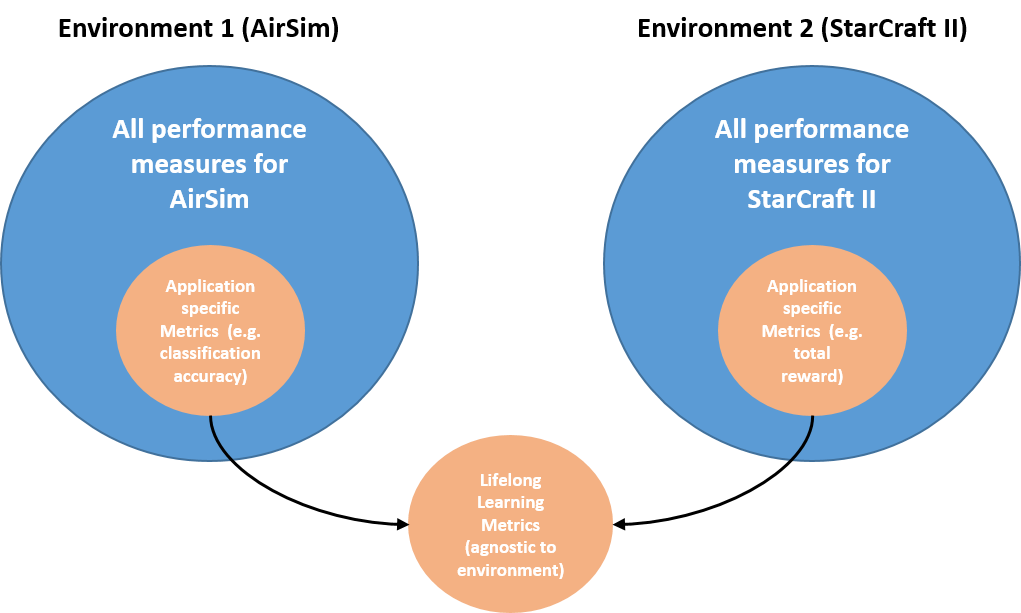}
    \caption{Environments such as AirSim or StarCraft generate many application-specific performance measures, such as classification accuracy, number of enemy units defeated, or total reward. Some subset of the values reported by the Environment is needed to calculate the Lifelong Learning Metrics (\cref{sec:metrics}), but it is not necessary to choose the \textit{same} application-specific measure for computing all of the \gls{LL} Metrics, since these measures are tracked over the course of the \gls{LL} system lifetime. For example, the number of enemy units defeated may be used to compute one metric, and total reward may be used to compute another. This allows a system to be evaluated in a flexible, environment-agnostic way. Figure adapted from~\cite{New2022lifelong}.}
\label{fig:performance-measures}
\end{figure}

As shown in \cref{fig:performance-measures}, an \gls{LL} system performing tasks in its environment as specified by the Evaluation Protocol will generate some number of application-specific measures. Each \gls{LX} -- the minimum amount of experience with a task that enables some learning activity on the part of the system -- is assumed to generate one or more scenario, domain, environment, and application-specific performance measures. A chosen subset of these application-specific measures is tracked and used to compute the \gls{LL} Metrics. It is important to note that a task's application-specific performance measures in a scenario will only be compared to the same task's same application-specific performance measures. For example, consider an \gls{LL} system that encounters two tasks $A$ and $B$. Before encountering task $B$, the system has a performance value for task $A$ of $P_{A,\mathrm{before}}$; after encountering task $B$, the system has a performance value for task $A$ of $P_{A,\mathrm{after}}$. Then, as defined in~\cref{bt}, we can assess how learning $B$ changes performance on $A$ with the \gls{BT} score:
$$\mathrm{BT}_{B\to A} = \frac{P_{A,\mathrm{after}}}{P_{A,\mathrm{before}}}.$$
There are no comparisons made between the performance values of $A$ and $B$ to compute these \gls{LL} Metrics, so there is correspondingly no need to choose only one application-specific measure to assess an \gls{LL} system's performance across all tasks. In order to summarize the \gls{LL} system's performance for each Metric in a scenario, we used mean aggregation, but other options are possible.

In the following section we discuss each \gls{LL} Condition, including the motivation for assessing it, the metrics associated with doing so, and the question that the metric attempts to address. At the end of each subsection, we provide \gls{LL} threshold values for the metrics associated with that \gls{LL} Condition.

\subsection{Continuous Learning Metrics}
A system demonstrating Continuous Learning  will consolidate new information to improve performance while coping with irrelevance, noise, and distribution shift. The \gls{LL} system needs to discover and adaptively select or ignore information that may be relevant or irrelevant. In particular, a Lifelong Learner must not be plagued by catastrophic forgetting, and performance must quickly recover when the agent is re-introduced to tasks whose performance may have degraded. While we have a metric to address whether a system has catastrophically forgotten task data, our attempt at formulating a metric to address whether a system recovers after a drop in performance was unsuccessful and is discussed more in Section \ref{sec:discussion}.

\subsubsection{Performance Maintenance (PM)}
A Lifelong Learner should be capable of maintaining performance on each of its tasks. \gls{PM} measures whether an \gls{LL} system catastrophically forgets a previously learned task and compares a system's performance when it first has the opportunity to learn a task to subsequent times experiencing the task. An important caveat here is that \gls{PM} does not measure absolute performance levels; rather, it measures a change in performance over the course of the system's lifetime. While there is some overlap between what \gls{PM} and \gls{BT} measure (~\cref{sec:transfer}), \gls{BT} compares a particular task's \glspl{EB} immediately before and after learning a new task, whereas \gls{PM} can be computed using any sequence of \glspl{EB}, independent of how many other tasks were learned between. 

\subsubsection{LL threshold value for Performance Maintenance}
The \gls{LL} threshold value for \gls{PM} is zero - this value indicates that, on average, there are no differences between initial and subsequent performances on a task. A positive value would indicate improvement over the course of a lifetime - a potential indicator of transfer. A negative value indicates forgetting. It is worth noting that this metric may be particularly sensitive to high variance in the application-specific measure ranges, since the metric computes a difference rather than use a ratio or a contrast. 

\begin{table}[h]
\centering
\begin{tabular}{|l|l|}
\hline
Case    &  Interpretation \\
\hline
PM $>$ 0 &  \begin{tabular}{@{}c@{}}(Demonstrates LL) that performance on task is getting \\better over lifetime; may be an indication of transfer.\end{tabular}\\
\hline
PM = 0 &  No forgetting; no additional learning.   \\
\hline
PM $<$ 0 & (Does not demonstrate LL) Indicates forgetting. \\
\hline
\end{tabular}
\caption{LL Threshold values for Performance Maintenance}
\label{tab:pm-ll-thresholds}
\end{table}

\subsection{Transfer and Adaptation Metrics}\label{sec:transfer}
One of the hallmark capabilities of a system capable of \gls{LL} is the ability to leverage experience on one task toward improving performance on another. Without assuming knowledge of the details of \textit{how} a system may accomplish this, we can measure progress toward this aim by computing both forward and backward transfer. At the very least, we expect that an \gls{LL} system will not exhibit catastrophic forgetting, where learning a new task interferes with performance of a previously learned task.

For this particular suite of metrics, \gls{FT} was formulated as a jumpstart measure as introduced by ~\cite{taylor2007crossdomain}, where performance changes were assessed at the beginning of a learning block, measuring whether the system got a ``jumpstart'' on a future task. We used this formulation for \gls{FT} for two primary reasons. First, the intention of these metrics was to be as domain-agnostic as possible, and addressing the nuance of how a learning curve changed could require a substantial amount of computational resources. Second, the preference was for a single value to express a system's performance for each of the metrics, where possible. Transfer has been defined differently by others, but a jumpstart measure enables evaluation of the beginning of a system's lifetime, which we felt was most appropriate given that we were assessing widely different systems. An important implication of this formulation to note is that for interpretability purposes, a forward transfer value is computed for only the first two tasks in a sequence.

\subsubsection{Forward Transfer (FT)}

\gls{FT} involves a system utilizing experience from prior, seen tasks to improve on a future, unseen task. Importantly, since a primary aim in developing these metrics is their application without consideration of task specifics, we compute \gls{FT} \textbf{only in the first instance of each task pair} as the ratio of the application-specific measure in an evaluation block before and after another task is learned. As formulated, this metric measures whether the \gls{LL} system leverages data from a previously learned task to learn a new task, and it requires the presence of Evaluation Blocks before and after each new task's first Learning Block in order to be computed. An important note about \gls{FT} is that order of the tasks is important. \gls{FT} may be present from Task A $\to$ B, but not Task B $\to$ A.

\subsubsection{Backward Transfer (BT)} \label{bt}
A system demonstrating \gls{BT} will use expertise in a new task to improve performance on a known task. Unlike \gls{FT}, which is only computed on the first instance of each task pair, \gls{BT} can be computed for each task after every \gls{LB}. This metric measures whether an \gls{LL} system leverages data from a new task to improve performance on a previously learned task, and it requires \glspl{EB} between each \gls{LB} to measure the performance after new tasks are learned. \gls{BT} is computed for each task where scenario structure allows.

\subsubsection{LL Thresholds for Forward and Backward Transfer}
\cref{tab:ft-bt-ll-thresholds} shows the \gls{LL} threshold values for both \gls{FT} and \gls{BT}. A value of 1 would demonstrate no change in task performance, meaning neither forgetting nor transfer, whereas values above or below 1 would indicate transfer and interference, respectively.

\begin{table}[h]
\centering
\begin{tabular}{|l|l|}
\hline
Case    &  Interpretation \\
\hline
BT / FT $>$ 1 &  (Demonstrates \gls{LL}) Indicates positive forward transfer.   \\
\hline
BT / FT = 1 &  No transfer or forgetting   \\
\hline
BT / FT $<$ 1 & (Does not demonstrate LL) Indicates interference. \\
\hline
\end{tabular}
\caption{LL Threshold values for Forward and Backward Transfer}
\label{tab:ft-bt-ll-thresholds}
\end{table}

\subsection{Scalability Metrics}

A fundamental capability for operationalized or deployable ML systems is the use of limited resources (e.g., memory, time) to accomplish or learn tasks in a scalable way.  We expect an \gls{LL} system to be able to sustain learning activity for arbitrarily long lifetimes including many tasks, though in practice,``arbitrarily long'' and ``many tasks'' are relative to typical operational timescales of the application domain. While there are several ways to assess the use of limited resources, one domain-agnostic methods for doing so (used by~\cite{Hayes2018NewParadigms}) is to compare the performance of an \gls{LL} system that is trying to learn many tasks to a \gls{STE} system that is learning just one task. The Sustainablity metrics assess essential components of \gls{LL} because it is useful to see if an \gls{LL} system is being outperformed by individual subsystems trained for each task. Scalability Metrics are an important component of system performance, in addition to being a proxy for task capacity.

\subsubsection{Performance Relative to a Single Task Expert (RP)}
An LL system with good \gls{RP} will perform well on each of its tasks when directly compared to the corresponding \gls{STE}, often leveraging data from other tasks to do so. As formulated, \gls{RP} measures how the performance of an \gls{LL} system compares to a non \gls{LL} system with comparable training. \gls{RP} is related to the Transfer metrics in that a system that exhibits strong \gls{FT} or \gls{BT} should benefit from these effects. However, \gls{RP} offers a more complete look at performance that combines all of the experience on a particular task and compares it to the performance of a \gls{STE} with a similar amount of experience.

\subsubsection{Sample Efficiency (SE)}
Lifelong Learners are expected to sustain learning over long periods of time. The rate of performance gain of a system is a part of scalability; a system that learns quickly is efficient with the amount of experience it is exposed to. As formulated, \gls{SE} describes the rate of task performance
gain with additional experience. This metric measures the performance gain of the \gls{LL} system by comparing the absolute level of performance (the ``saturation value'') achieved by the \gls{LL} system and the number of learning experiences required to get there with the corresponding \gls{STE} values. 

\subsubsection{LL Thresholds for Relative Performance and Sample Efficiency}

Determining threshold values for \gls{LL} is more nuanced for the Scalability metrics. Ideally, we want the performance of an \gls{LL} system to match or exceed that of an \gls{STE}, as reflected in the determination of the LL thresholds in \cref{tab:rp-se-ll-thresholds}.

\begin{table}[h]
\centering
\begin{tabular}{|p{0.2\linewidth} | p{0.75\linewidth}|}
\hline
Case    &  Interpretation \\
\hline
RP / SE $>$ 1 & (Demonstrates LL) Indicates Performance / Performance Gain above level of STE \\
\hline
RP / SE = 1 & Indicates Performance / Performance Gain exactly at level of STE \\
\hline
RP / SE $<$ 1 & (Does not demonstrate LL) Indicates Performance / Performance Gain below level of STE \\
\hline
\end{tabular}
\caption{\gls{LL} threshold values for \gls{RP} and \gls{SE}. STE indicates Single Task Expert}
\label{tab:rp-se-ll-thresholds}
\end{table}

\section{Case Studies with Lifelong Learning Systems}\label{sec:case_studies}

In this section, we examine five System Group case studies, all of which exercised the suite of \gls{LL} Metrics. These metrics were computed on \gls{LL} systems developed during the \gls{L2M} Program using various techniques and in different environments, as shown in \cref{tab:ll-systems}. Over the course of \gls{L2M}, we conducted multiple system evaluations, which are denoted by M12, M15, and M18. Each of the following subsections contains a brief overview of the corresponding \gls{LL} system developed by each SG team, a description of the tasks used in each of the environments (summarized in \cref{tab:envs-and-tasks}), and a discussion of results and insights provided by the Metrics. For more details regarding the specific implementation of these systems and/or the results they generate, please see the referenced published work.

\subsection{System Group UPenn - AIHabitat} \label{sec:sg-upenn}

\subsubsection{System Overview}
This section describes a case study on the development of the \gls{LL} system led by SG-UPenn, a modular system that performs both classification and \gls{RL} tasks in realistic service robot settings.  The core of the system, which integrates factorized models (\glspl{DFCNN} for supervised learning~\citep{lee2019learning} and \gls{LPGFTW} for \gls{RL}~\citep{Mendez2020}), is divided into separate classification and \gls{RL} pipelines, with the perception-action loop of a mobile robot. 
The system includes additional optional modules that can be combined with the core classification and \gls{RL} pipelines, including a task-agnostic feature meta-learning module using \gls{METAKFO}~\citep{arnold2021maml}, intrinsic motivation via meta-learned intrinsic reward functions~\citep{zheng2020can}, an alternative core \gls{RL} algorithm based on the \gls{A2C} algorithm~\citep{mnih2016asynchronous}, a self-supervised exploration module based on active visual mapping for robot navigation~\citep{ramakrishnan2020occupancy}, and a \gls{MDP}-based curriculum learning module~\citep{narvekar2020curriculum}.  These components can be turned on and off depending on the problem domain, and characterizing their effects through the set of \gls{LL} Metrics proposed in this paper was a focus of the experimentation discussed in this case study. The task settings and select experimental results for the two pipelines are described below.

\subsubsection{Classification Experimental Context}

\textbf{Classification.}  Lifelong classification experiments were carried out by SG-UPenn over data sets collected by simulated agents performing random walks through household environments in the AI Habitat simulator~\citep{habitat19iccv} using the Matterport 3D data set~\citep{Chang2017Matterport3D}, resulting in realistic observations for household service robots derived from real world sensor data.  All experiments were conducted over a fixed curriculum of object classification tasks, where each task required a mobile agent to classify a set of objects taken from an object superclass, e.g. classifying \{\texttt{chair}, \texttt{sofa}, \texttt{cushion}, \texttt{misc\_seating}\} from the superclass \texttt{seating\_furniture}.

\subsubsection{Classification Experimental Results}

\begin{table}[h]
\centering
\resizebox{\textwidth}{!}{
\begin{tabular}{|l|c|c|c|c|c|}
\hline
Configuration & PM & FT & BT & RP & SE \\
\hline
DF-CNN & $-0.44  \pm 1.12$ & $1.01  \pm 0.09 $ & $ 0.99  \pm 0.02 $ & $1.94  \pm 0.26 $ & $1.61  \pm 0.12 $ \\
META-KFO & $-20.81  \pm 15.22 $ & $1.00  \pm 0.00 $ & $0.91  \pm 0.07 $ & $2.38  \pm 0.40 $ & $3.40  \pm 0.46 $ \\
\hline
\end{tabular}
}
\caption{Select SG-UPenn classification experiment results.  All metrics show mean  $\pm$ standard deviation.}
\label{tab:upenn-cl-results}
\end{table}

This case study focuses on a specific classification experiment for which the proposed set of \gls{LL} Metrics was particularly informative.  The goal of this experiment was to determine the differences in performance between factorized classification models and meta-learned classification models in a lifelong supervised learning setting.  To explore this, SG-UPenn ran the same set of Lifelong classification experiments over two configurations of the system:  the (factorized) \gls{DFCNN} core classification pipeline and the (meta-learned) META-KFO module.  The results (Table~\ref{tab:upenn-cl-results}) show that, while both approaches show good \gls{LL} performance, META-KFO provides faster learning (higher \gls{SE}) whereas the DF-CNN provides more stable learning through better catastrophic forgetting mitigation (higher \gls{PM} and \gls{BT}, with lower standard deviations).  As such, SG-UPenn prioritized future development of the DF-CNN pipeline due to the stability afforded by the factorized method.

\subsubsection{Reinforcement Learning Experimental Context}

\textbf{Reinforcement Learning.}  Lifelong \gls{RL} experiments were carried out in the AI Habitat simulator using the Matterport 3D data set.  All experiments were conducted over a fixed curriculum of object search tasks in the form of ``find a given object (e.g. a chair, a cabinet, a sink, or a plant) in a given household environment (e.g. an apartment or a town house).''  The agents observed RGB images from a head-mounted camera, and their actions were direct control commands.  %The agents used image features pre-trained on the ImageNet dataset, but had no knowledge of the house layouts, decor, or target objects until deployment in the lifelong learning scenario.

\subsubsection{Reinforcement Learning Experimental Results}

\begin{table}[h]
\centering
\resizebox{\textwidth}{!}{
\begin{tabular}{|l|c|c|c|c|c|}
\hline
Configuration & PM & FT & BT & RP & SE \\
\hline
M12 & $-60.1 \pm 21.5$ & $0.89 \pm -0.80$ & $1.2 \pm 1.56$ & $0.75 \pm 0.07$ & $0.66 \pm 0.27$ \\
M15 & $-14.0 \pm 20.5$ & $1.95 \pm 0.97$ & $1.19 \pm 0.16$ & $0.75 \pm 0.06$ & $1.88 \pm 1.96$ \\
M18 & $4.4 \pm 11.3$ & $3.11 \pm 2.36$ & $1.11 \pm 0.07$ & $0.88 \pm 0.03$ & $0.83 \pm $0.03 \\
\hline
\end{tabular}
}
\caption{Select UPenn System Group reinforcement learning experiment results.  All metrics show mean $\pm$ standard deviation.}
\label{tab:upenn-rl-results}
\end{table}

The first \gls{RL} experiment (M12) hypothesized that intrinsic motivation would improve \gls{FT}, \gls{RP}, and \gls{SE} in \gls{LL} settings, making it an effective mechanism for knowledge reuse in lifelong \gls{RL}.  To test this hypothesis, SG-UPenn used the intrinsic motivation module combined with the core \gls{A2C} \gls{RL} algorithm.  The results did not support this hypothesis, instead showing that intrinsic motivation is not an effective mechanism for lifelong learning, as shown in the M12 column of Table~\ref{tab:upenn-rl-results}.  The main issue identified was that the system was highly susceptible to catastrophic forgetting, as evidenced by the particularly low \gls{PM} score.  To overcome this problem, SG-UPenn focused system development on factorized methods instead, which are specifically designed to mitigate catastrophic forgetting.

The next set of \gls{RL} experiments (M15) focused on evaluating the effectiveness of the factorized \gls{LPGFTW} algorithm in the realistic Habitat/Matterport environment.  This system configuration used the core \gls{LPGFTW} algorithm with no additional modules.  The results show significant improvement compared to the intrinsic motivation pipeline across all of the Lifelong Learning Metrics, with the exception of comparable \gls{RP}.  SG-UPenn notes that while the \gls{PM} score was still negative, it is significantly higher than the intrinsic motivation pipeline, which shows increased mitigation of catastrophic forgetting.  SG-UPenn continued to develop the \gls{LPGFTW}-based system with additional network architecture search and hyperparameter tuning that targeted the \gls{PM} metric.  Shown in the M18 column of Table~\ref{tab:upenn-rl-results}, this resulted in significant improvements to both \gls{PM} and \gls{FT}.  Contrary to the experimental results in the original \gls{LPGFTW} paper~\citep{Mendez2020}, there is still relatively low performance with respect to single task experts (i.e. in the \gls{RP} and \gls{SE} metrics).  SG-UPenn hypothesizes that this performance drop is due to the increased challenge of learning in high fidelity environments, and the higher task complexity that such environments entail.

\subsection{System Group Teledyne - AirSim} \label{sec:sg-teledyne}

\subsubsection{System Overview}
This section describes a case study on the development of the \gls{LL} system led by SG-Teledyne. It consists of six key components, the core of which is the \gls{UML}~\citep{Brna2019} algorithm. \gls{UML} enables adaptation and learning in response to multiple types of uncertainty. Inspired by mechanisms of neuromodulation, \gls{UML} compares its internal hypotheses against expectations and adapts its behavior based on the level of mismatch. Under high uncertainty, it re-configures itself and re-evaluates its inputs, allowing robust operation in noisy environments or in the presence of new conditions. Under low uncertainty, the algorithm can more confidently engage in long-term adaptation to learn new tasks or tune its knowledge base. Because uncertainty serves to gate learning and the type of adaptation in the system, it can prevent catastrophic forgetting and promote behaviorally-relevant adaptation. Furthermore, under very high uncertainty conditions, \gls{UML} protects existing knowledge to allow one-shot learning of novel information. Finally, the algorithm can use its internal measures of uncertainty to actively seek new information to optimize learning and resource utilization ~\citep{Brown2022}. A limitation of \gls{UML} is that it requires a robust representation of its inputs. Nonetheless, it has proven to work well when using the output layer of deep neural networks trained on datasets such as ImageNet~\citep{deng2009imagenet} or COCO~\citep{Lin2014coco}. Another limitation is that it learns to recognize tasks by the difference in the context of each task. Therefore, there is a requirement that each task possesses a sufficiently different context.

\subsubsection{Experimental Context}

The \gls{UML} algorithm has been evaluated in multiple \gls{ML} domains, including classification~\citep{nMNIST2017}), embodied agents~\citep{Brna2019},~\citep{Brown2022}, and reinforcement learning. 
Under DARPA L2M, \gls{UML} was evaluated using an embodied agent. Data was generated using AirSim~\citep{Airsim2017} in a custom Unreal Engine 4 environment. The classification tasks were split into two ``Asset Groups'' loosely corresponding to notional municipal interest groups: EMA (Emergency Management) vehicles and DOT (Department of Transportation) traffic control assets (e.g., stop signs, traffic lights, etc.). Each asset group contained 2-3 individual classes of objects. The classification problems associated with each asset group formed tasks, and variants of those tasks were generated using different environmental conditions (e.g., time of day). 

Experiments were conducted on permutations in ordering of these task variants, with a full evaluation across tasks being conducted after each exposure to a task.

\subsubsection{Experimental Results}\label{sec:teledyne_results}

Table~\ref{tab:tdy-cl-results} shows aggregate results across all such runs generated using the SG-Teledyne system, which matched or exceeded the \gls{LL} threshold value in all 5 metrics across the collected runs. These metrics enabled us to evaluate the performance of individual components in the system and their impact on \gls{LL} capabilities. In an ablation experiment, \gls{TDY} showed that the memory consolidation technique in one of the system components (C5) was responsible for a significant gain in \gls{FT}, but at the expense of \gls{PM}, while other metrics remained relatively constant. These metrics enabled a deeper analysis and more complete understanding of the the impact of this component as it relates to the \gls{LL} characteristics.

\begin{table}[h]
\centering
\resizebox{\textwidth}{!}{
\begin{tabular}{|l|c|c|c|c|c|}
\hline
Configuration & PM & FT & BT & RP & SE \\
\hline
TDY UML Agent & $0.56 \pm 0.98$ & $11.69 \pm 0.47$ & $1.00 \pm 0.01$ & $1.03 \pm 0.04$ & $2.74 \pm 1.70$ \\
TDY C5 Ablation & $1.68 \pm 0.36$ & $10.47 \pm 0.23$ & $1.02 \pm 0.02$ & $1.01 \pm 0.03$ & $2.33 \pm 0.74$ \\
\hline
\end{tabular}
}
\caption{Selected SG-Teledyne experiment results.  All metrics show mean $\pm$ standard deviation. The baseline agent is shown in the \gls{TDY} \gls{UML} Agent row, and a selected ablation experiment is shown in the \gls{TDY} C5 Ablation row. The metrics enabled us to understand the effects of the ablation study on specific \gls{LL} characteristics.}
\label{tab:tdy-cl-results}
\end{table}

\subsection{System Group HRL - CARLA} \label{sec:sg-hrl}

\subsubsection{System Overview}
This section describes a case study on the \gls{STELLAR}, the \gls{LL} system developed by SG-HRL. \gls{STELLAR} is a general-purpose, scalable autonomous system capable of continual online \gls{RL} that is applicable to a wide range of autonomous system applications, including autonomous ground vehicles (both on-road and off-road), autonomous undersea vehicles, and autonomous aircraft, among others. It consists of a deep convolutional encoder that feeds into an actor-critic network and is trained using Proximal Policy Optimization \citep{Schulman2017PPO}. Importantly, \gls{STELLAR} integrated 11 innovative components that solve different challenges and requirements for \gls{LL}. It employed Sliced Cramer Preservation (SCP) \citep{kolouri2020sliced}, or the sketched  version of it (SCP++) \citep{Li2021SCP++}, and Complex Synapse Optimizer \citep{Fusi2016Metaplasticity} to overcome catastrophic forgetting of old tasks; Self-Preserving World Model \citep{Ketz2019WorldModels} and Context-Skill Model \citep{Tutum2021CSM} for backward transfer to old tasks as well as forward transfer to their variants; Neuromodulated Attention \citep{Zou2020Attention} for rapid performance recovery when an old task repeats; Modulated Hebbian Network \citep{Ladosz2021MOHN} and Plastic Neuromodulated Network \citep{Ese2021PNN} for rapid adaptation to new tasks; Reflexive Adaptation \citep{Maguire2021Reflex} and Meta-Learned Instinct Network \citep{Grbic2021Instinct} to safely adapt to new tasks; and Probabilistic Program Neurogenesis \citep{Martin2019PPN} to scale up the learning of new tasks during fielded operation. More details on the precise effect of each of these components are beyond the scope of this paper; however, this case study outlines how the integrated system dynamics demonstrated \gls{LL} using the proposed metrics, and how these metrics shaped the advancement of the SG-HRL system.
\subsubsection{Experimental Context}
\gls{STELLAR} was evaluated within the CARLA driving simulator~\citep{Dosovitskiy17CARLA} in both the Condensed and Dispersed \gls{LL} Scenarios (described in~\cref{sec:ll_scenarios}), which were each based on three tasks with two variants per task. The agent was required to drive safely from one point to another within a designated lane (either correct or opposite) in traffic. It was given positive rewards in each time step (every 50 ms) for distance traveled towards the destination and increasing speed within the designated lane. It was given negative rewards for distance traveled away from the destination and decreasing speed within the designated lane, as well as any collision. A given episode was terminated when the destination was reached, a maximum number of time steps had elapsed, or there was any collision. SG-HRL employed two vehicle models (Audi TT [car], Kawasaki Ninja [motorcycle]) with built-in differences in physical parameters such as for the body (e.g., mass, drag coefficient) and wheels (e.g., friction, damping rate, maximum steering angle, radius). The vehicle models also differed in camera orientation (0$^{\circ}$ yaw for car vs. 45$^{\circ}$ yaw for motorcycle).

The same architecture as the \gls{STELLAR} systems was used to train the \glspl{STE} to saturation, thereby characterizing the ability of the \glspl{STE} to learn each task. SG-HRL collected 10 \gls{STE} runs per task, which were all initialized with the same “ready-to-deploy” state as the \gls{STELLAR} system. 

\subsubsection{Experimental Results}\label{sec:hrl_experiment}
Given that the \gls{STELLAR} system integrates the 11 components listed above with the specific intent to achieve various \gls{LL} capabilities, SG-HRL expected the metrics to reveal such properties of the system. Indeed in both Condensed and Dispersed scenarios, the \gls{STELLAR} system exceeded the threshold for \gls{LL} for 4 of the 5 metrics, with only a non-catastrophic degradation in \gls{PM} of old tasks through the lifetimes (Table~\ref{tab:HRL_Results}).

\begin{table}[h]
\centering
\resizebox{\textwidth}{!}{
\begin{tabular}{|l|c|c|c|c|c|}
\hline
Configuration & PM & FT & BT & RP & SE \\
\hline
Condensed (n=33) & $-0.24 \pm 5.73$ & $10.02 \pm 4.92$ & $1.19 \pm 0.26$ & $2.49 \pm 1.31$ & $10.02 \pm 13.88$ \\
Dispersed (n=30) & $-2.21 \pm 3.16$ & $10.71 \pm 2.78$ & $1.10 \pm 0.15$ & $1.85 \pm 0.71$ & $6.25 \pm 3.12$ \\
\hline
\end{tabular}
}
\caption{\gls{LL} performance of the STELLAR system in the Condensed and Dispersed scenarios within the CARLA driving simulator. Mean $\pm$ standard deviation values for each metric are shown across n=33 and n=30 lifetimes, respectively, comprising random permutations of tasks and variants.}
\label{tab:HRL_Results}
\end{table}

Further, as shown in ~\cref{tab:HRL_Results}, SG-HRL found that the performance was not significantly different between the Condensed and Dispersed scenarios. 
% \alex{probably point to~\cref{tab:HRL_Results_b} here?}
However, all the \gls{LL} Metrics were numerically lower for the Dispersed scenario, with the decrements being significant at $\alpha = 0.1$ for two metrics; namely, \gls{FT} ($p$ = 0.089, Mann-Whitney U Test) and \gls{RP} ($p$ = 0.038, Mann-Whitney U Test). Potential explanations for the across-the-board numerical decrements in the metrics include: the increased cost of switching among tasks in the Dispersed scenario, greater interference from other tasks in the intervals between learning blocks for a given task, or the lack of any dependence of the strength of the consolidation mechanisms (SCP++, Self-Preserving World Model) on the performance levels acquired in the preceding learning blocks. In the Dispersed scenario, task performances in earlier learning blocks are not expected to be high due to shorter durations. In this case, strong preservation of sub-optimal task representations would interfere with subsequent learning blocks. Thus, the hyperparameters that control the degree of preservation should be reduced to improve all the \gls{LL} Metrics.  

\begin{table}[h]
\centering
\resizebox{\textwidth}{!}{
\begin{tabular}{|l|c|c|c|c|c|}
\hline
Configuration & PM & FT & BT & RP & SE \\
\hline
Dispersed & $-2.73 \pm 2.71$ & $9.96 \pm 2.16$ & $1.15 \pm 0.18$ & $1.57 \pm 0.49$ & $7.07 \pm 3.44$ \\
Reduced SCP++ stiffness & $0.26 \pm 3.84$ & $9.52 \pm 2.97$ & $1.27 \pm 0.29$ & $2.07 \pm 0.44$ & $3.23 \pm 1.42$ \\
\hline
\end{tabular}
}
\caption{Summary of the effects of reducing SCP++ stiffness on the Dispersed scenario for the \gls{STELLAR} system. Dispersed results (n=15) represent a subset of data shown in \ref{tab:HRL_Results}. SCP++ stiffness reduction (n=15) results from matched lifetimes. All results show mean $\pm$ standard deviation.}
\label{tab:HRL_ablation}
\end{table}

The STELLAR system requires considerable analysis to assess how each component contributes to various \gls{LL} capabilities. This case study represents one such analysis to illustrate the impact on the metrics. SG-HRL hypothesized that stronger consolidation mechanisms would reduce \gls{LL} in the Dispersed scenario which, unlike the Condensed scenario, has task repetitions. SG-HRL also predicted that strong consolidation of sub-optimal task representations after each task would negatively impact subsequent learning blocks. Data was collected for the Dispersed scenario with the SCP++ stiffness coefficient reduced to 10\% of the nominal value (Table~\ref{tab:HRL_ablation}). As expected, SCP++ stiffness reduction resulted in improvements in 3 of the 5 metrics; namely, \gls{PM} (from -2.73 to 0.26), \gls{BT} by about 10\%, and \gls{RP} by about 30\%. But the manipulation also caused decrements in the other 2 metrics; namely, \gls{FT} by about 4\% and \gls{SE} by about 50\%. Of these effects, the improvement in \gls{RP} ($p$ = 0.022, Wilcoxon Signed Rank Test) and the decrement in \gls{SE} ($p$=0.0026, Wilcoxon Signed Rank Test) were statistically significant, and the improvement in \gls{PM} ($p$=0.055, Wilcoxon Signed Rank Test) was significant at $\alpha = 0.1$. More work will be needed to understand the dynamics of \gls{LL} for task repetitions in the context of the multi-component STELLAR system. It may be the case that the degree of consolidation (structural regularization, interleaving of explicit/generative replays) should be further contingent on task learning, and SG-HRL anticipates testing this in the future.

\subsection{System Group Argonne - L2Explorer} \label{sec:sg-argonne}

\subsubsection{System Overview}

This section describes a case study on the development of the \gls{LL} system led by SG-Argonne. The system's design was inspired by the brains of insects and other small animals with the motivation of developing systems capable of \gls{LL} that can operate effectively at the edge~\citep{ayg_insect}.

In particular, it focuses on the use of: 1) modulatory learning and processing, which control how information is processed, as well as when and where learning takes place~\citep{anurag_modnet}; 2) metaplasticity models, which modulate synaptic plasticity rules that keep either a memory or an internal state in order to preserve useful information~\citep{VandeVen_2019}; 3) broadly trained representations, which apply transfer learning to minimize what the system needs to learn during deployment, and 4) structural sparsity, which minimizes the impact of forgetting by curtailing gradient propagation in stochastic gradient descent methods~\citep{Sandeep_2020}.

In the context of \gls{RL}, Argonne adapted these principles to propose two types of algorithms. First, they proposed a lifelong deep Q learning algorithm~\citep{Mnih2013atari} aimed at solving problems where a consistent policy is learned across a series of independent tasks without specific task labels. Second, they proposed a lifelong cross entropy algorithm, which applies to situations involving short, potentially contradictory tasks, where no prior information is available that would lead to accurate and consistent computations of the value of each state. For the case of deep Q learning, Argonne's system realizes short term and long term memory buffers by implementing periodic shuffling. The size of the buffers is kept within the length of a single task.

\subsubsection{Experimental Context}
Over the course of the project, SG-Argonne worked in two different environments. The first and more complex environment was L2Explorer~\citep{Johnson2022L2Explorer}, a first-person point of view environment built on top of the Unity engine~\citep{Juliana2018unity} that allows the creation of tasks involving open-world exploration. Argonne designed a series of tasks emphasizing different aspects of a complex policy involving target identification and selection, navigation through obstacles, navigation towards landmarks, and foraging objects while avoiding hazards.
The same tasks were implemented in Roundworld, a lightweight, first-person point of view environment developed by Argonne that comprises a simpler set of objects and visual inputs, allowing us to evaluate the algorithm across two different environments.

\subsubsection{Experimental Results}

~\cref{tab:anl-M18M21} shows the performance of the deep Q learning algorithms in the two different environments. In both cases there is a consistent evidence of both forward and backward transfer across tasks in the proposed scenario. One of the characteristic aspects of these environments is their task variability, both by design and driven by the open world nature of the environments. In the context of \gls{RL}, this leads to large fluctuations in the values of \gls{PM} and \gls{BT} for both environments, with standard deviations more than one order of magnitude higher than those typically observed in image classification scenarios. On the other hand, both scenarios show values of \gls{FT}, \gls{RP}, and \gls{SE} that are consistent with the presence of \gls{LL} behaviors.

\begin{table}[htp]
    \centering
    \resizebox{\textwidth}{!}{
    \begin{tabular}{|c|c|c|c|c|c|c|c|}
        \hline
         Environment & Scenario & Agent & PM & FT & BT & RP & SE \\
         \hline
  
 L2Explorer   &  Condensed & M18 & $-4$ $\pm 11$ & $4.6$ $ \pm 1.5$ & $2.3$ $ \pm 1.6$ & $1.2$ $ \pm 0.6$ & $1.2$ $ \pm 0.6$ \\
  
 Roundworld   &  Condensed & M18 & $15$ $\pm 10$ & $4.2$ $ \pm 1.6$ & $2.7$ $ \pm 2.1$ & $5$ $ \pm 3.4$ & $5.8$ $ \pm 1$\\

\hline
    \end{tabular}
    }
    \caption{Evaluation of the lifelong deep-Q learning algorithm in two different environments with varying complexity levels.}
    \label{tab:anl-M18M21}
\end{table}

Having access to different metrics allows for deeper insight into variations in the system's performance. Overall, the results obtained point to a complex picture in which the same Lifelong Learning system can exhibit different behavior depending on how well it can transfer information during its lifetime. However, further studies are needed in order to fully explore how the behavior of the agent depends on task sequence and its ability to effectively transfer relevant policies across tasks.

\subsection{System Group SRI - StarCraft II} \label{sec:sg-sri}
\subsubsection{System Overview}

This section describes a case study on the development of the \gls{LL} system led by SG-SRI. The system is targeted at real-time strategy games where task change occurs naturally and throughout game play. For example, a competent Starcraft-2 (SC2) player is able to adapt their tactics to different enemy units. This section applies lifelong \gls{RL} techniques to micromanagement tasks in SC2. This case study shows that the proposed metrics (a) validate that the negative effects of task drift are mitigated, (b) drive algorithm development to improve metrics, and (c) provide insights into software integration of multiple continual learners.

Components of the SG-SRI system \citep{sur2022system,daniels2022model} include: (i) WATCH \citep{faber2021watch,faber2022lifewatch}, a Wasserstein-based statistical changepoint detection that detects changes in the  environment; (ii) Self-Taught Associative Memory (STAM) \citep{stam_ijcai}, to generate feature maps from RGB images in a continually updated manner; (iii) Danger detection, using the continual learner deep streaming linear discriminant analysis (DeepSLDA) \citep{hayes2020lifelong}; (iv) Compression, using the REMIND algorithm \citep{hayes2020remind} that uses Product Quantization (PQ); and (v) Sleep phase,  implemented using the Eigentask framework \citep{raghavan_lifelong_2020}.

\subsubsection{Experimental Context}
The tasks are defined using different SC2 maps called ``minigames`` \citep{StarCraft}. The system is evaluated on the minigames of \textit{DefeatRoaches}, \textit{DefeatZerglingsAndBanelings} and \textit{CollectMineralShards}. To each task, SG-SRI added a variant of the task that spawns two groups of enemies on each side of the map, creating a total of 3 tasks and 2 variants each. In the case of \textit{Collect}, the variant has fog enabled (partial observability). SG-SRI notes that combat related tasks (\textit{Defeat}) are most similar to each other (due to their reward structure) and represent 4 out of 6 tasks, so high forward transfer (jumpstart) is expected even for the single task learner.

\subsubsection{Experimental Results}

Table~\ref{tab:algo-M121518} shows evolution of the Eigentask algorithm driven by the proposed \gls{LL} Metrics, with the current version of the system (denoted M18) achieving the criteria of lifelong learning in all but one metric (PM) in the condensed scenario and achieving the criteria of \gls{LL} in several metrics for the alternating scenario. These versions, denoted as M12, M15, M18, correspond to the evaluations performed under \gls{L2M}. These versions primarily differ in the generative replay architecture. The M12 model connects the autoencoders and policies one after another, whereas M15 uses a two-headed architecture using a common latent space and M18 uses hidden replay. In both scenarios, the metrics show that the M18 version that uses hidden replay is a significant improvement. Of note, the reported metrics have significantly lower variance with the M18 model compared to the M15 and M12 versions for the condensed scenario.

\begin{table}[htp]
    \centering
    \resizebox{\textwidth}{!}{
    \begin{tabular}{|c|c|c|c|c|c|c|}
        \hline
         Scenario & Agent & PM & FT & BT & RP & SE \\
         \hline
         %& M12 & & & & \\
         & M12 & $-3.70$ $ \pm 2.5$ & $1.15$ $ \pm 0.06$ & $1.00$ $ \pm 0.13$ & $0.91$ $ \pm 0.13$ & $12.22$ $ \pm 4.97$\\
         Condensed & M15 & $-5.68$ $ \pm 5.04$ & $1.42$ $ \pm 0.25$ & $1.14$ $ \pm 0.28$ & $1.18$ $ \pm 0.19$ & $19.37$ $ \pm 5.76$\\
         & M18 & $-3.05$ $ \pm 1.76$ & $1.42$ $ \pm 0.11$ & $1.0$ $ \pm 0.03$ & $1.17$ $ \pm 0.11$  & $16.18$ $ \pm 5.19$\\
         \hline
         & M12 & $-7.44$ $ \pm 6.19$ & $1.18$ $ \pm 0.67$ & $0.88$ $ \pm 0.14$ & $0.80$ $ \pm 0.14$  & $4.74$ $ \pm 2.27$ \\
         %Pairwise & M15 & $-5.4$ ($ \pm 11.61$) & $1.13$ ($ \pm 0.57$) & $0.84$ ($ \pm 0.32$) & $1.06$ ($ \pm 0.23$) & $13.78$ ($ \pm 7.24$)\\
         %& M18 & $-3.88$ ($ \pm 7.68$) & $1.54$ ($ \pm 0.31$) & $0.86$ ($ \pm 0.21$) & $1.04$ ($ \pm 0.15$)  & $120.08$ ($ \pm 483.42$)\\
         %\hline
         % & M12 & $-7.44$ ($ \pm 6.19$) & $1.18$ ($ \pm 0.67$) & $0.88$ ($ \pm 0.14$) & $0.8$ ($ \pm 0.14$) & $4.74$ ($ \pm 2.35$)\\
         Alternating & M15 & $-8.82$ $ \pm 7.95$ & $1.13$ $ \pm 0.57$ & $0.80$ $ \pm 0.19$ & $0.90$ $ \pm 0.11$ & $7.11$ $ \pm 3.52$\\
         %& M18 & & & & &\\
        & M18 & $-6.13$ $ \pm 7.31$ & $1.85$ $ \pm 1.38$ & $0.87$ $ \pm 0.27$ & $0.91$ $ \pm 0.13$  & $5.89$ $ \pm 3.19$\\

         \hline
    \end{tabular}
    }
    \caption{Evolution of the SRI-led \gls{LL} system guided by the proposed metrics. Pairwise scenarios are averaged over 12 lifetimes.}
    \label{tab:algo-M121518}
\end{table}
 
To study the effect that change detection and compression had on the overall performance of the \gls{LL} system, SG-SRI performed an ablation experiment against the baseline Eigentask component in two different scenario types. \gls{PM} and \gls{BT} values are compared in Table~\ref{tab:sri-system}, showing that triggering the sleep phase by statistical changepoint detection results in significantly higher PM compared to triggering it by a hand-coded schedule. This demonstrates the importance of task detection in \gls{LL} systems in the task-agnostic setting and also shows that the compression of wake phase observations results in significantly higher PM. This ablation experiment demonstrates how the metrics shed insight on the impact of various system components during the development of the SG-SRI \gls{LL} system.

\begin{table}[htp]
    \centering
    \resizebox{\textwidth}{!}{%
    \begin{tabular}{|c|c|c|c|c|}
         \hline 
         Agent & \multicolumn{2}{|c|}{Performance Maintenance} & \multicolumn{2}{|c|}{Backward Transfer}  \\
         \hline 
         & Condensed & Pairwise & Condensed & Pairwise \\
         \hline
         Single Task Learner (STL) & $-3.41$ ($ \pm 1.7$) & $-8.2$ ($ \pm 6.54$) & $1.17$ ($ \pm 0.29$) & $0.85$ ($ \pm 0.21$) \\
         Eigentask (M15) & $-5.68$ ($ \pm 2.13$) & $-5.40$ ($ \pm 4.9$) & $1.14$ ($ \pm 0.12$) & $0.84$ ($ \pm 0.12$) \\
         Eigentask + Change detection & $-0.53$ ($ \pm 4.49$) & $-1.93$ ($ \pm 5.46$) & $1.02$ ($ \pm 0.33$) & $1.08$ ($ \pm 0.28$) \\
         Eigentask + Compression & $-3.67$ ($ \pm 3.92$) & $-2.23$ ($ \pm 2.33$) & $1.13$ ($ \pm 0.42$) & $0.93$ ($ \pm 0.22$)\\
         \hline 
    \end{tabular}
    }
    \caption{Ablations comparing system components on Performance Maintenance and Backward Transfer. The standard error is mentioned in parenthesis ($\pm$). Other metrics are omitted for brevity.}
    \label{tab:sri-system}
\end{table}

\subsection{Summary of Case Studies of Systems Demonstrating LL}

In this section we have reviewed five System Group case studies, all of which operated in different environments and employed different algorithms.  Each of them used the suite of \gls{LL} Metrics to inform their system development and evaluate whether their systems demonstrated the Conditions of Lifelong Learning in various experiments. In \cref{tab:metric-t-tests} we see that across all of the System Groups, the Lifelong Learning thresholds were met or exceeded for 52 out of 90 metrics, with Performance Maintenance only meeting the LL Threshold values in 3 of the 18 configurations compared to 13 configurations for Forward Transfer. This is unsurprising given that Performance Maintenance and Forward Transfer represent different aspects of the well-known performance trade-off between stability and plasticity, which we discuss further in \cref{sec:discussion}.
\begin{table}
\scalebox{0.85} {
\begin{tabular}{|p{0.12\linewidth}|p{0.23\linewidth}|p{0.15\linewidth}|p{0.15\linewidth}|p{0.15\linewidth}|p{0.15\linewidth}|p{0.15\linewidth}|}
    \hline
        SG & Config & PM & FT & BT & RP  & SE \\ \hline
        UPenn & DF-CNN & $1.20 \cdot 10^{-2}$ & $2.67 \cdot 10^{-1}$ & $1.58 \cdot 10^{-2}$ & \textless{}$10^{-6}$ & \textless{}$10^{-6}$ \\
        & META-KFO & \textless{}$10^{-6}$ & $1.00$ & \textless{}$10^{-6}$ & \textless{}$10^{-6}$ & \textless{}$10^{-6}$ \\
        & RL M12 & $5.65 \cdot 10^{-3}$ & $4.01 \cdot 10^{-1}$ & $4.07 \cdot 10^{-1}$ &  $2.83 \cdot 10^{-3}$ & $4.31 \cdot 10^{-2}$ \\
        & RL M15 & $1.86 \cdot 10^{-2}$ & $3.02 \cdot 10^{-3}$ & $8.65 \cdot 10^{-4}$ & \textless{}$10^{-6}$ & $7.35 \cdot 10^{-2}$ \\
        & RL M18 & $1.71 \cdot 10^{-1}$ & $7.02 \cdot 10^{-3}$ & $1.98 \cdot 10^{-4}$ &  \textless{}$10^{-6}$ & \textless{}$10^{-6}$ \\ \hline
        Teledyne & C5 Ablated & $3.68 \cdot 10^{-2}$ & \textless{}$10^{-6}$ & $2.42 \cdot 10^{-1}$ & $1.01 \cdot 10^{-2}$ & $2.29 \cdot 10^{-3}$ \\
        & UML & $8.35 \cdot 10^{-3}$ & \textless{}$10^{-6}$ & $4.59 \cdot 10^{-3}$ & $4.96 \cdot 10^{-2}$ & $4.76 \cdot 10^{-3}$ \\ \hline
        HRL & Condensed & $5.90 \cdot 10^{-1}$ & \textless{}$10^{-6}$ &$1.74 \cdot 10^{-4}$ &  \textless{}$10^{-6}$ & $4.32 \cdot 10^{-4}$ \\ 
        & Dispersed &$1.00$ & \textless{}$10^{-6}$ & $1.53 \cdot 10^{-4}$ & \textless{}$10^{-6}$ & \textless{}$10^{-6}$ \\
        & SCP Ablation & $4.00 \cdot 10^{-1}$ & \textless{}$10^{-6}$ & $2.00 \cdot 10^{-3}$ &  \textless{}$10^{-6}$ & $2.03 \cdot 10^{-5}$ \\ \hline
        Argonne & L2Explorer & $1.07 \cdot 10^{-1}$ & \textless{}$10^{-6}$ &$6.31 \cdot 10^{-3}$ &  $1.26 \cdot 10^{-1}$ & $1.26 \cdot 10^{-1}$ \\
        & Roundworld & $1.48 \cdot 10^{-4}$ & $1.20 \cdot 10^{-5}$ &$8.57 \cdot 10^{-3}$ &  $9.17 \cdot 10^{-4}$ & \textless{}$10^{-6}$ \\  \hline
        SRI & M12 Condensed & $1.00$ & $2.28 \cdot 10^{-6}$ & $5.46 \cdot 10^{-1}$ & $9.81 \cdot 10^{-1}$ & $4.09 \cdot 10^{-5}$ \\
        & M15 Condensed & $1.00$ & \textless{}$10^{-6}$ & $1.10 \cdot 10^{-2}$ & $6.88 \cdot 10^{-5}$ & \textless{}$10^{-6}$ \\
        & M18 Condensed & $1.00$ & \textless{}$10^{-6}$ & $2.73 \cdot 10^{-1}$ & $2.66 \cdot 10^{-5}$ & \textless{}$10^{-6}$ \\
        & M12 Alternating & $1.00$ & $1.55 \cdot 10^{-1}$ & $9.98 \cdot 10^{-1}$ & $1.00$ & $1.25 \cdot 10^{-5}$ \\
        & M15 Alternating & $1.00$ & $1.32 \cdot 10^{-1}$ & $1.00$ & $1.00$ & \textless{}$10^{-6}$ \\
        & M18 Alternating & $9.93 \cdot 10^{-1}$ & $2.82 \cdot 10^{-2}$ & $9.80 \cdot 10^{-1}$ & $9.86 \cdot 10^{-1}$ & $1.76 \cdot 10^{-4}$ \\ \hline

\end{tabular}
} % End scale box
    \caption{P value results of a one-tailed t-test to determine whether the value is significantly greater than the LL Threshold value for that metric; t values are provided in \cref{tab:metric-t-tests-t-values} of \ref{sec:t-test-vals}. The \gls{LL} threshold values were met or exceeded for 45 out of 85 metrics.
    Note that the UPenn META-KFO system was designed to speed up the rate of adapting to a new task, but this does not happen until data for that task is seen, leading to unchanged task values and a standard deviation of zero for a jumpstart formulation of FT.}
\label{tab:metric-t-tests}
\end{table}

\glsresetall

\section{Discussion}\label{sec:discussion}

In this work, we have proposed and investigated a suite of domain- and technique- agnostic metrics to enable a systems-level development approach for evaluating Lifelong Learning systems. Such an approach is critical to supporting the multi-objective nature of \gls{LL} system development, especially because increasingly complex solutions are required to advance the state of the art towards \gls{LL}. 
A strength of our approach is that it simultaneously considers and quantifies varied capabilities of \gls{LL} systems, rather than focusing on any single aspect of performance. 
By using the full suite of metrics to evaluate the System Group case studies, we were able to identify and study the performance trade-offs inherent to \gls{LL}.
Next, we discuss known performance trade-offs seen with these metrics, propose a new trade-off, and make recommendations for creating additional metrics for future investigations based on the accomplishments of the \gls{L2M} program.

\subsection{LL Performance Trade-offs}

We have argued that \gls{LL} is complex and cannot be characterized by a single scalar value. This has motivated our development of a suite of metrics. 

Designing an \gls{LL} system must consider the following trade-offs:
\begin{enumerate}
    \item Stability vs. Plasticity: Should a system stably maintain all information it has encountered up to some point, even if that results in less flexibility to adapt to changes?
    \item Optimal Performance vs. Computational Cost: Should a system be optimized for maximum performance, even if that comes at a high computational cost?
    \item Sample Efficient vs. Robust Learning: Should a system prioritize a fast performance gain, even if it is less robust to noise or changes in the environment?
\end{enumerate}

The most widely discussed trade-off in \gls{LL} literature is the relationship between Stability, where a system has reliable or low-variance performance, and Plasticity, where a system is flexible and adaptable to changes (see, e.g., discussion in~\cite{Mermillod2013stabilityplasticity,Grossberg1988stability}). \gls{PM} is a measure of stability, since it assesses how well a system retains task knowledge gained over the course of its lifetime; \gls{FT} is a measure of plasticity, as it assesses how well a system can apply knowledge from one task to another. In some cases, like the stiffness parameter experiment examined in SG-HRL's case study (see~\cref{tab:HRL_ablation}), there is an explicit parameter that can be tuned, depending on the needs of the particular application, to prioritize reliability or flexibility. This results in somewhat expected behavior changes.
In other cases, the trade-off is seen as a byproduct of targeting improvements in transfer, like in SG-Teledyne's addition of a memory consolidation component (see ~\cref{tab:tdy-cl-results}), which manages the system's stored knowledge. This addition caused marked improvement in \gls{FT} -- a measure of Plasticity -- but at the cost of \gls{PM}, a measure of Stability.

\indent It is understood that \gls{LL} systems operating in diverse environments will have varied design considerations; the availability or restriction of computational resources is one such factor. This can result in an intentional decision to choose system components that are less performant but cheaper computationally. While this discussion surfaces in the literature, particularly with regard to deployment considerations, we chose not to measure the computational resource expenditure for these evaluations. Instead, we allowed system groups to make their own assessments of progress in their domain. Even if an \gls{LL} system is initially very computationally intensive, it may be possible to develop a more efficient system in the future. In non-\gls{LL}, existing techniques for managing model complexity include: model distillation~\citep{Hinton2015distillation,Gou2021distillation}, intelligently-designed model scaling strategies~\citep{Tan19EfficientNet}, and investigations of broad scaling phenomena~\citep{Kaplan2020nlpscaling}. These approaches could potentially be extended to LL; in~\cite{hayes2020remind}, SG-SRI built on a technique called progress \& compress~\citep{schwarz2018progress}. We see the addition of a metric to standardize the measurement of resource utilization as an excellent extension of this suite, and we summarize some initial efforts in this area in ~\ref{sec:cc}. We collect our comments, observations and recommendations for the design and use of such a metric in \cref{sec:metrics_observations}.

We hypothesize that, as more progress is made to develop LL systems, more of these system design/performance trade-offs will be discovered. One trade-off that we observed in the SG-UPenn case study (\cref{sec:sg-upenn}) was between sample-efficient and robust learning. The system's robustness to task or parameter changes was measured using the \gls{PM} metric, and efficiency was measured via the \gls{SE} metric. We can imagine a situation where a system may have an extremely robust representation of a wide range of tasks -- along the lines of a subject matter expert for a particular problem space -- but perhaps amassing that knowledge required significant training data and time. Conversely, a system may demonstrate aptitude for rapid mastery, but lack the broader experience to capably handle the details or nuance of edge cases. 

The trade-off, then, may be that in some circumstances, optimizing for robustness comes at the cost of learning efficiency and vice versa.  This goal is particularly relevant in data-poor contexts or where the cost of training is high; both of these apply in many robotic applications (like the SG-UPenn service robot setting). The \gls{LL} system the SG-UPenn team built to address these challenges includes modularized components and factorized models, an approach that is well-suited to these conditions. Correspondingly, we see that when modifications were made between M15 and M18 systems to target gains in \gls{PM} (\cref{tab:upenn-rl-results}), the resulting M18 results improved in \gls{PM}, but at the cost of a lower Sample Efficiency. This demonstrates a consequence of the opposing aims of Robust and Sample Efficient learning. We imagine that this trade-off may not be applicable to problems with low-cost or abundant training data. However, it is apparent in this particular example, because SG-UPenn's system design is intended for eventual transfer to service robot settings.

\subsection{General Considerations for Formulation and Use of  Metrics}\label{sec:metrics_observations}

One of the challenges of measuring \gls{LL} performance is evaluating over the space of possible task sequences. 
\noindent Because these tasks may require orthogonal skills, it is an immense challenge to quantify a priori what ideal or even ‘good’ performance looks like for such a sequence. 
\noindent The standard we chose for determining thresholds for \gls{LL}, which can certainly over-penalize an \gls{LL} system, was perfection -- perfect transfer between tasks and perfect memory of a task over the entire duration of a scenario. Over the course of the agent's lifetime, any interference, forgetting, or performance not equal to or better than an STE was considered to be below the threshold for \gls{LL}. 
\noindent Meeting this threshold for all lifelong learning conditions is likely to be difficult in real-world conditions. Determining an appropriate upper bound for performance on a sequence of tasks is a fundamental challenge -- one that requires leveraging information like task difficulty and task similarity (and thus task transferability) -- and was out of scope for this work.
\noindent Below we outline some specific recommendations for metric design, some of which pose particularly unique challenges in the \gls{LL} domain.

\begin{enumerate}
\item \textbf{Do not design metrics that rely on idealized performance curves} \\
\noindent Despite knowing that we lack the ability to quantify what good performance is for a given sequence of tasks, there were some unanticipated difficulties in using the metrics related to some key assumptions about the nature and behavior of LL systems:
\begin{itemize}
\item \textit{Assumption 1: Learning a particular sequence of tasks is possible.} \\
\noindent When we develop metrics to evaluate any machine learning system, we are often doing so based on an implicit assumption that a task is learnable by the system or, at least, that the system is capable of demonstrating some performance gain over 

the course of its \glspl{LX}. In the absence of baseline approaches on the same sequence of tasks to compare to, we may not even be able to say whether a sequence of tasks is learnable at all without running a cost-prohibitive number of experiments. In fact, the idea of learnability in the Lifelong Learning context has only recently been investigated in works such as ~\cite{Geisa2021theory}, who explores the relationship between weak and strong learnability for both in-distribution (i.e. non-LL) and out of distribution problems. As the theory of learnability for Lifelong Learning is still developing, we must design our metrics acknowledging the potential for systems to demonstrate no learning on some tasks and, importantly, address whether or not those runs should be considered in computing the LL metrics. The results shown in \cref{sec:case_studies} included all runs, independent of whether tasks demonstrated learning.
\item \textit{Assumption 2: In learning a sequence of tasks, performance on a previously learned task may drop, but it can and will ``bounce back'' when the task is shown later.} \\
\noindent This assumption drove the design of the Performance Recovery metric, which in theory was designed to measure whether an \gls{LL} system's performance recovers after a change is introduced to its environment. To compute Performance Recovery, we first calculated the number of learning experiences the system required in order to get back to the previously attained value after a drop (\textit{recovery time}), and computed the change in this number of experiences over the course of the system's lifetime (i.e., fitted a line to the \textit{recovery times} and computed the slope of the line). The idea was that a system demonstrating LL would adapt more quickly to changes as it amassed more experience. 

Of note, Performance Recovery could only be assessed for scenarios with many task repetitions. The use of this metric proved to be problematic, in particular because some systems would fail to ``bounce back'' sufficiently. This dependency of final system performance on initial \glspl{LX} has been observed in the broader deep \gls{RL} space~\citep{Nikishin2022primacybias}, where it was aligned with the concept of ``primacy bias'' from human cognition studies~\citep{marshall1972primacybias}. Beyond this binary challenge of a system returning to previous performance or not; given the variability in the application-specific measures, it also remained difficult to discern when performance has actually ``bounced back'' and to what should the new performance be compared, and how should we handle noise in these measurements? \cite{Dror2019Dominance} recommends the use of the Almost Stochastic Dominance test to mitigate the variability issue we faced, but we were unable to implement this due to the computational expense associated with this analysis.

\item \textit{Assumption 3: We can identify when a task has been ``learned,'' or at least, when the system performance has saturated.}\\
\noindent Computing whether a system's performance has saturated (and to what value) is not straightforward, in part due to the heteroskedastic nature of the learning curves. There is unpredictability to system learning, and coupling this with noisy learning makes this computation even more of a challenge. In addition, the notion of ``saturation'' may be ill-defined, particularly when the distribution of an environment within a learning block is nonstationary.
In the case of this suite of metrics, Sample Efficiency explicitly relies on the computation of a saturation value, and Performance Maintenance compares an average of the most recent training performance to future evaluation performances -- with the implicit assumption that a system has reached a stable, if not maximal, performance value at the end of a learning block.
\end{itemize}
In light of these challenges, we recommend designing an assessment -- even a simple performance threshold specific to an environment -- to determine whether a system has learned and to lend insight into computed metric values. \\

\item \textbf{Do not avoid metrics that measure overlapping concepts.}\\

Due to some similarities in their formulation, we expected some of the metrics (e.g. PM and BT, SE and RP) to be strongly correlated. In practice, we found only weak positive correlations between those two metric pairs, as shown in Table~\ref{tab:metrics_correlations}. We also found that SE and PM were weakly negatively correlated, which supports our discovery of a performance trade-off between these two metrics. The strongest correlation across the metrics was between Forward Transfer and Relative Performance at $\rho$ = 0.45. This correlation makes sense in retrospect - a system which excels at Forward Transfer is likely to require fewer learning experiences for a task (and thus have a higher RP score) if it can benefit from another task's learning experiences as well. 
\noindent Even in the case of the most correlated metrics, it was critical to have both measures since they offer an assessment of a different LL condition and add an additional perspective on assessing the whole system's performance. \\

\item \textbf{Design metrics with clear interpretations based on the LL thresholds.}\\
In light of the difficulty of determining an upper bound for an agent's performance on a sequence of tasks, we made two intentional choices when formulating and interpreting the metrics. In their formulation, the LL thresholds for the metrics are clearly delineated, giving a straightforward interpretation - values above the threshold demonstrate the corresponding condition of LL, and values below do not. This was extremely useful for interpreting values and determining whether a system demonstrated lifelong learning. Though we formulated the metrics such that larger scores are better, this binary interpretation of each of the metrics allowed for a systems level analysis of performance rather than a specific focus on any one measure. 
\\
\item \textbf{Compare performance to an STE when possible} \\
\noindent Overall, our most robust measure of LL was the metric that baselined performance to a single task expert - Relative Performance. 
Relative Performance offered insight into the question of whether a system is demonstrating an improvement over previous attempts to do lifelong learning versus simply assessing whether a system demonstrates lifelong learning. This comparison to a benchmark can also be used to indicate progress over previous approaches -- similar to an ablation experiment -- but functions primarily as a proxy for establishing an upper bound of performance on any given task.
\\
\\
\item \textbf{Be cautious in estimating properties of data from noisy reward function distributions}\\
As discussed in Section~\ref{sec:evaluation_protocol}, Reinforcement Learning systems can be especially noisy. To remediate some issues that arise from computing values on noisy data, we preprocessed the data by smoothing it and shifting the range to exclude zero to avoid the vanishing denominator issue.
\noindent In light of the noise intrinsic to these environments described by  \citep{Agarwal2021deep}, we recommend keeping metric formulations simple.  We also recommend being especially wary of second order metrics, like Performance Recovery, where noise can be compounded to the point of ineffectiveness. We hope to reformulate Performance Recovery in the future.
\\

\item \textbf{Be cautious about application specific metric ranges and their potential effect on ratios}
In initial formulations of Forward and Backward Transfer, we compared the performance before and after relevant task learning as a standard ratio under the assumption that it was unlikely for a system to achieve zero (or very small values) as an application specific measure of performance. This assumption, unfortunately, did not hold to be true. To address robustness issues that arose in those circumstances from an infinitesimal denominator, we added an alternative formulation of both forward and backward transfer using the contrast function:
$$\mathrm{Contrast}(a, b) = \frac{a-b}{a+b}$$
where $a$ and $b$ represent a particular task performance either before or after learning a new task. While qualitatively similar to the ratio function, $\mathrm{Ratio}(a,b)= \frac{a}{b}$, contrasts differ in that they are defined when $b = 0$. This ensures they are well-defined in situations where application-specific measures are or approach zero; while the stability is a benefit, it can be less intuitive and therefore more complicated to interpret. Due to this difficulty in interpretation, we reported the ratio values in Section~\ref{sec:case_studies}.
\end{enumerate}

\begin{table}[]
    \centering
    \begin{tabular}{|c|c|l|l|}
        \hline
        Metric 1            &  Metric 2             & Spearman Corr.        & $p$-value\\\hline\hline
        Perf. Maintenance   &  Forward Transfer     & 0.06                  & 0.60  \\
                            &  Backward Transfer    & 0.33                  & 0.003 \\
                            &  Relative Perf.       & -0.20                 & 0.07 \\
                            &  Sample Efficiency    & -0.25                 & 0.03 \\\hline
        Forward Transfer    &  Backward Transfer    & -0.09                 & 0.44 \\
                            &  Relative Perf.       & 0.45                  & 0.00003\\
                            &  Sample Efficiency    & 0.01                  & 0.93\\\hline
        Backward Transfer   &  Relative Perf.       & -0.15                 & 0.19\\
                            &  Sample Efficiency    & -0.16                 & 0.14\\\hline
        Relative Perf.      &  Sample Efficiency    & 0.36                  & 0.001\\\hline
    \end{tabular}
    \caption{Correlation analysis of values of different metrics. Despite expecting strong correlations between PM and BT as well as SE and RP, these metric pairs were only weakly correlated. We found that SE and PM were weakly negatively correlated, which supports our discovery of a performance trade-off between these two metrics.}
    \label{tab:metrics_correlations}
\end{table}

\section{Conclusion}
In this work, we argued that evaluating advances in Lifelong Learning is a complex challenge that requires a systems approach to assessing performance and quantifying trade-offs, especially since there are currently no universally accepted metrics for Lifelong Learning. We presented the Conditions that an \gls{LL} system should demonstrate as a Lifelong Learner, and developed a suite of metrics to assess those Conditions. We outlined a method for calculating the metrics in a scenario, domain, environment, and task-agnostic fashion to characterize capabilities of LL systems.
We demonstrated the use of the suite of metrics via five case studies that used varied environments, illustrating the strengths and weaknesses of each system using the metrics. We discussed the quantification of three key performance trade-offs present in the development of many LL systems, and made recommendations for future metric development for LL systems.

Though the field of \gls{LL} is nascent, methods and metrics for comprehensive evaluation are a critical piece in realizing a future with operationalized \gls{ML} systems. As these \gls{LL} systems increase in complexity to address current limitations, the challenge of evaluating performance and identifying strengths and weaknesses will become both more difficult and more crucial, especially in domains such as military operations or healthcare. Using a consistent suite of metrics for evaluation of complex systems in a domain- and technique-agnostic way enables a complete tracking of progress across the entire field of \gls{LL}.

Many challenges remain in evaluating \gls{LL} systems, including extending the computation of metrics across all lifetimes of a system, adding additional metrics to consistently quantify the computational cost trade-off, and formulating metrics that measure or account for relationships or properties of various tasks. Our suite of metrics provides a basis for extensions that can address these and other newly-discovered gaps.

\section{Acknowledgements}

Primary development of this work was funded by the DARPA Lifelong Learning Machines (L2M) Program. The authors would like to thank the performers in this program, particularly the members of the Definitions and Metrics Working Groups, who helped to develop and refine these LL conditions and metrics. Thanks also goes to the DARPA SETAs for this program--Conrad Bell, Robert McFarland, Rebecca McFarland, and Ben Epstein. Useful feedback on improving the manuscript was provided by Alice Jackson, Kiran Karra, Jared Markowitz, and Christopher Ratto.

\section{Disclaimer}
The views, opinions, and/or findings expressed are those of the author(s) and should not be interpreted as representing the official views or policies of the Department of Defense or the U.S. Government.

\appendix

\glsresetall

\section{Terminology}\label{sec:terminology}

\begin{table}[!ht]
    \centering
    \begin{tabular}{|p{0.2\linewidth} | p{0.8\linewidth}|}
    \hline
        Term & Definition  \\ \hline
        Task & Some non-trivial capability that the agent must learn, and on which performance is directly measured. A task should have parameters for stochastic and structured variation (sufficient to pose a challenging learning problem), and should have some notion of generalization. For example, in the domain of sports, ``Tennis'' and “Badminton” would be tasks. \\ \hline
        Task Variants & Variants of a task are substantially different versions of a task - different enough to pose a significant learning challenge, and outside of the range of stochastic variation. For example, ``Tennis on grass court during day'' and ``Tennis on clay court at night'' may be considered variants. \\ \hline
        % Conversely, ``Tennis with angle of sunlight at 90 degrees'' and ``Tennis with angle of sunlight at 95 degrees'' would not be considered variants as the changes fall within range of stochastic variation. 
        Task Instance & A specific occurrence of a task that an agent encounters. In the sports domain, ``Tennis'' is a task, and an instance of Tennis would be a single game of tennis, on a specific kind of court, at a specific time of day and weather, with specific initial conditions, and so on. \\ \hline
        \Gls{LX} & A minimum amount of experience with a task that enables some learning activity on the part of the agent. A task instance can be a single \gls{LX}, or it might consist of multiple \glspl{LX}.  \\ \hline
        \Gls{EX} & A minimum amount of experience with a task that enables some demonstration of learned activity on the part of the agent. During an \gls{EX}, the LL system is being evaluated at a ``frozen'' state and no weight updates are allowed.
        \\ \hline
        Block & A sequence of Experiences for a single task/variant. May be a \gls{LB} or an \gls{EB}
        \\ \hline
        Lifetime & A sequence of \glspl{LB} and \glspl{EB} encountered by the agent once it is deployed. A lifetime starts with the agent in a ``Ready-to-deploy'' state.  \\ \hline
        Lifelong Learning Scenario & A scenario characterizes a single lifetime for an agent. It consists of a set of tasks (or task variants), any related parameterization, and optionally, specifications on how the tasks should be sequenced in the lifetime.  \\ \hline
        Evaluation Protocol & An evaluation protocol is a complete specification for getting statistically reliable \gls{LL} metrics. It consists of a specification of pre-deployment training, one or more scenarios, and how multiple lifetimes (runs) are generated for each scenario. \\ \hline
    \end{tabular}
    \label{tab:terminology}
\end{table}

\glsresetall

\section{Supplementary Information about Scenarios}\label{sec:supplementary_scenario_info}

\subsection{Condensed and Dispersed Scenarios}
\begin{figure}[htbp!]
  \centering
  \includegraphics[width=1.0\linewidth]{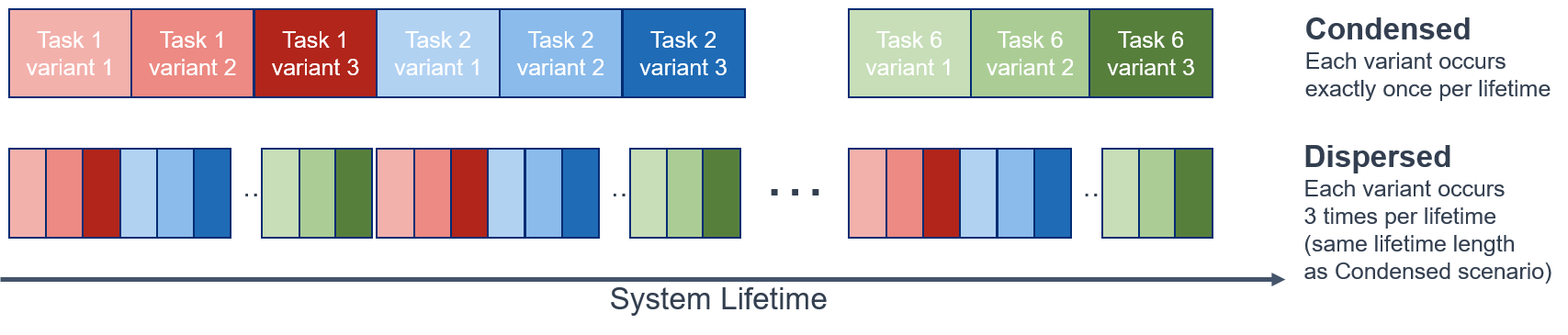}
    \caption{Illustration of Condensed and Dispersed Scenario Types introduced in \cref{sec:ll_scenarios}}
\label{fig:scenario-types-appendix}
\end{figure}

We consider two key types of evaluation scenarios. Both consist of an interleaving sequence of \glspl{LB} and \glspl{EB}. In the former, the \gls{LL} system encounters \glspl{LX} from a specific task and improves itself. In the latter, the \gls{LL} system encounters \glspl{EX} and is tested on how well it has mastered tasks. Beyond the two types here, many other variations are also devisable.

The condensed scenario assesses how well a system can retain performance on a wide variety of tasks. In it, \glspl{LB} for a given task variant occur only once in the scenario, and \glspl{LB} are chosen to be sufficiently long for the system to attain mastery on that block's task. 

In contrast, the dispersed scenario evaluates how well a system performs when the tasks it is exposed to change frequently. In this scenario, there are three ``superblocks'' (defined as a single permutation of task variants with shorter learning blocks, typically 1/3 the length of an \gls{LB} in a condensed scenario). A given task variant occurs exactly once during each superblock and each superblock uses a different random permutation of task variants.

\subsection{Example evaluation scenario}\label{sec:example_scenario}

In~\cref{tab:example_tasks}, we show how tasks and task variants can be defined for two environments--SplitMNIST~\citep{Zenke2017si,Shin2017dgr,nguyen2018variational}, and CARLA~\citep{Dosovitskiy17CARLA} environments, and in~\cref{tab:example_metrics}, we define the application-specific measures that assess \gls{LL} system performance on these tasks. Our framework of \glspl{LB} and \glspl{EB} is sufficiently general that we can represent two diverse scenario structures (condensed and dispersed scenarios), as well as two types of learning problems--classification for SplitMNIST and reinforcement learning for CARLA. Task variants can be defined by random (e.g., random brightness perturbations for Variant-2 of SplitMNIST's Task-1) or deterministic (e.g., fixed rotations for Variant-2 or SplitMNIST's Task-2) transformations. In addition, experiences can be subsampled from a finite dataset (SplitMNIST) or from a more complex generator (CARLA). If desired, similar tasks can use different application-specific measures (e.g., SplitMNIST's tasks using both \gls{ACC}~\citep{lopez2017gradient} and $\Omega_{all}$~\citep{Hayes2018NewParadigms}).

\begin{table}[htp!]
    \centering
    \resizebox{\textwidth}{!}{%
    \begin{tabular}{|p{0.1\textwidth}||p{0.45\textwidth}|p{0.45\textwidth}|}
    \hline
              &  SplitMNIST   &   CARLA\\\hline\hline
    Task-1    &  
    % SplitMNIST
    Classify images as being either \texttt{0} or \texttt{1}
    \begin{itemize}
        \item One \gls{LX} is a minibatch of sixteen images sampled from a training set
        \item One \gls{EX} is a minibatch of sixteen images sampled from a test set
        \item Variant-1: Images are left unaltered
        \item Variant-2: Images have their brightness randomly perturbed
        \item Variant-3: Images have their contrasts randomly perturbed
    \end{itemize}
    &
    % CARLA
    Task-1: Navigate from one point to another
    \begin{itemize}
        \item One \gls{LX} or \gls{EX} is one end-to-end navigation sequence
        \item Variant-1: There is little traffic
        \item Variant-2: There is heavy traffic
        \item Variant-3: Navigation sequences take place at nighttime
    \end{itemize}
    \\\hline
    Task-2
    &
    % SplitMNIST
    Classify images as being either \texttt{1} or \texttt{2}
    \begin{itemize}
        \item One \gls{LX} is a minibatch of sixteen images sampled from a training set
        \item One \gls{EX} is a minibatch of sixteen images sampled from a test set
        \item Variant-1: Images are left unaltered
        \item Variant-2: Images are rotated 90$^{\circ}$
        \item Variant-3: Images are rotated 270$^{\circ}$
    \end{itemize}
    &
    % CARLA
    Follow a sedan for a specified period of time
    \begin{itemize}
        \item One \gls{LX} or \gls{EX} is one end-to-end navigation sequence
        \item Variant-1: It is raining during navigation sequences
        \item Variant-2: The vehicle to be followed drives very quickly
        \item Variant-3: The vehicle to be followed is a semi-truck
    \end{itemize}
    \\\hline
    \end{tabular}
    }
    \caption{An example of how to construct two tasks and associated variants from the SplitMNIST and CARLA environments.}
    \label{tab:example_tasks}
\end{table}

\begin{table}[htp!]
    \centering
    \resizebox{\textwidth}{!}{%
    \begin{tabular}{|p{0.2\textwidth}||p{0.35\textwidth}|p{0.45\textwidth}|}
    \hline
              &  SplitMNIST   &   CARLA\\\hline\hline
Application-specific measures
    &
    % SplitMNIST
    \begin{itemize}
        \item Task-1: \gls{ACC}~\citep{lopez2017gradient}
        \item Task-2: $\Omega_{all}$~\citep{Hayes2018NewParadigms}
    \end{itemize}
    &
    % CARLA
    \begin{itemize}
        \item Task-1: Total travel time, penalized by unsafe driving
        \item Task-2: Average distance to target vehicle during the navigation sequence, penalized by unsafe driving
    \end{itemize}
    \\\hline
    \end{tabular}
    }
    \caption{An example of how to construct application-specific measures for tasks from the SplitMNIST and CARLA environments.}
    \label{tab:example_metrics}
\end{table}

\section{Additional details on metrics}\label{sec:additional_metrics_details}

\subsection{Notation for describing metrics and blocks}\label{sec:block_notation}

\begin{figure}
    \centering
    \includegraphics[width=\linewidth]{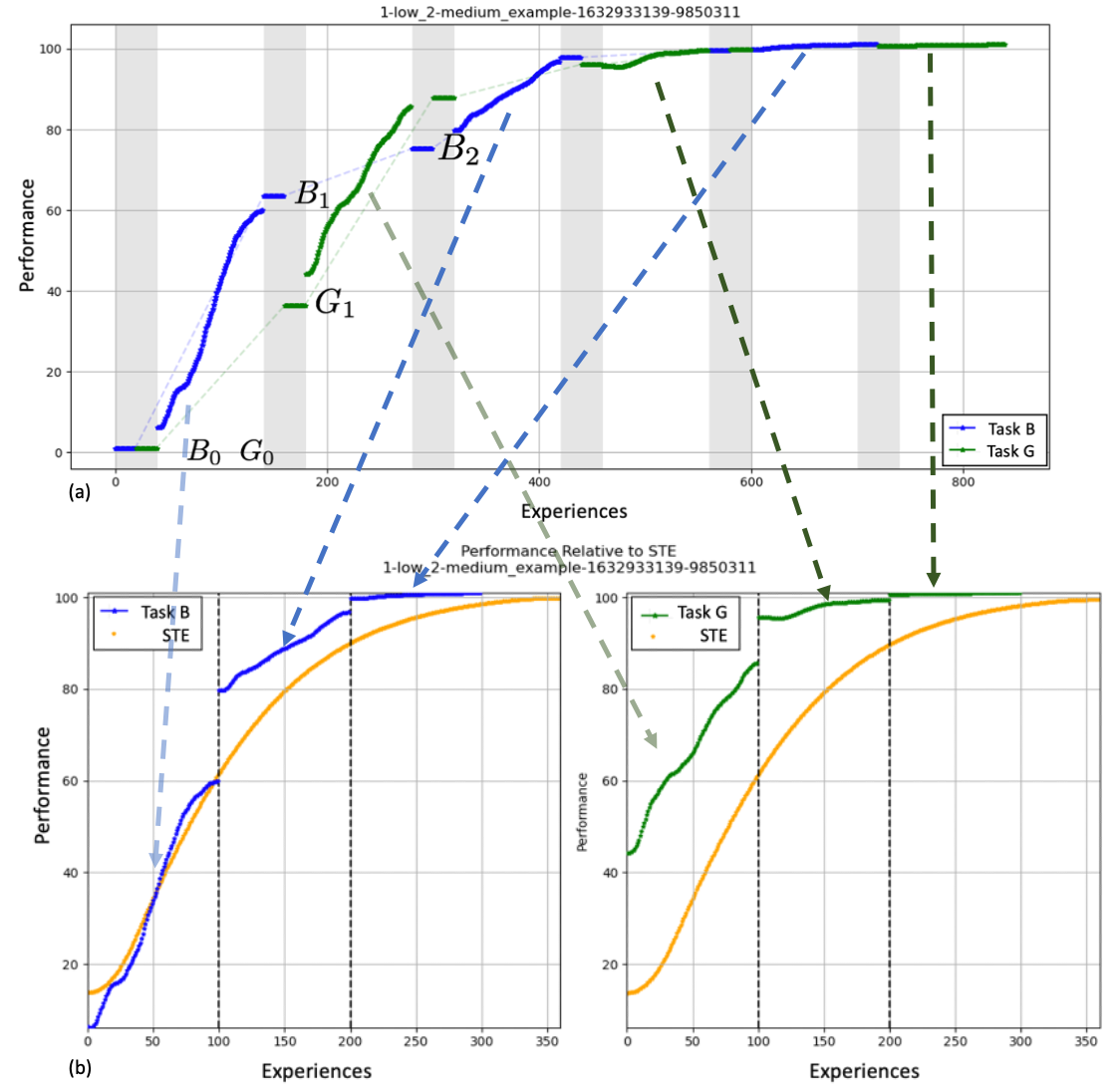}
    \caption{A notional lifetime containing two tasks, blue (B) and green (G).
    (a) The tasks alternate, and both are tested during evaluation blocks. The $y$-axis shows the agent's performance on tasks at each point during its lifetime. The $x$-axis counts the experiences of the lifetime. White shading corresponds to Learning Blocks, and grey shading corresponds to Evaluation Blocks.\\
    (b) Comparing the LL agent to single-task experts for the blue and green tasks (orange). Learning Blocks from the full lifetime are grouped by task and stitched together to form a task-specific learning curve.\\
    Figure adapted from~\cite{New2022lifelong}.}
    \label{fig:lifetime_fig}
\end{figure}

We introduce a compact set of notations to describe LL agent lifetimes and the quantities they output, illustrated in Figure~\ref{fig:lifetime_fig}.a. In general, a lifetime consists of $N$ Learning Blocks. During each learning block $n$, the agent is exposed to experiences from a single task $t(n)$ drawn from some larger set of possible tasks $\mathcal{T}$. Tasks may reoccur within a lifetime, or they may appear only once or not at all. After each Learning Block, an Evaluation Block occurs in which the agent is tested on all tasks in $\mathcal{T}$.

A Block consists of a sequence of (Learning or Evaluation) Experiences, and each Experience generates a single task-specific metric (e.g., a classification accuracy, reward function value, or binary outcome). These values must be preprocessed prior to calculation of LL metrics -- we recommend following the procedure described in Appendix A of \cite{New2022lifelong}, which is available in~\cite{Nguyen2022metrics}.

Ultimately, each Task $t$'s performance in Learning Block $n$ is summarized by a sequence of values $P_L(n,t) = (P_L(n,t,1),...,P_L(n,t,\ell(n)))$, and each Task $t$'s performance in the Evaluation Block after Learning Block $n$ is summarized by a scalar $P_E(n, t)$. Lifetimes are assumed to start with an Evaluation Block, yielding initial performance scores $P_E(0, t)$ for all $t \in \mathcal{T}$.

Baseline performance on a Task may be assessed by training a Single-Task Expert and an LL agent exposed to only one task. Relative Performance and Sample Efficiency metrics compare Learning Block performance of LL agents to STEs. We use $P_{STE}(n,t) = (P_{STE}(n,t,1), ..., P_{STE}(n,t,\ell(n)))$ to denote the performance in LX $\ell$ of the $n$th Learning Block of an STE trained on task $t$.

\subsection{Metric Formulations} \label{sec:metrics_formulations}

In this section, we present pseudo-code implementations of each of the metrics described in~\cref{sec:metrics}. Our transfer metrics (\cref{alg:ft} and~\cref{alg:bt}) use Ratios, but Contrasts may also be used in their place (see discussion in~\cref{sec:transfer} and~\cref{sec:metrics_observations}).

Our algorithms for metrics that consider data from single-task experts --  Relative Performance (\cref{alg:rp}) and Sample Efficiency (\cref{alg:se}) -- consider a simplified setting. Specifically, we assume that (1) for a given task, we have data from only a single STE, and (2) for a given task, the learning block lengths are the same across LL agents and STEs. The \texttt{l2metrics} package~\citep{Nguyen2022metrics} offers options for handling data when these assumptions are violated.

\begin{algorithm}
\caption{Calculation of Forward Transfer}\label{alg:ft}
\begin{algorithmic}
\Require Task set $\mathcal{T}$
\Require Evaluation Block Performances $\{P_E(n,t)\}$ for $n=0,...,N$, $t\in\mathcal{T}$
\Ensure $ForwardTransfer$
\vspace{0.2cm}
\State $FTs = LearnedTasks = LearnedTaskPairs = \emptyset$
\For{Learning Blocks $n=1,...,N$}
\If{$t(n)\not\in LearnedTasks$}
\State $LearnedTasks \gets LearnedTasks \cup \{t(n)\}$
\For{Tasks $t \in \mathcal{T}\setminus LearnedTasks$}
\If{$(t(n), t)\not\in LearnedTaskPairs$}
\State $(t(n), t)\gets LearnedTaskPairs\cup \{(t(n), t)\}$
\State $P_n, P_t = P_E(n,t), P_E(n-1, t)$
\State $FTs \gets FTs \cup \{\mathrm{Contrast}(P_n, P_t)\}$
\EndIf
\EndFor
\EndIf
\EndFor
\State $ForwardTransfer \gets \mathrm{mean}\{FTs\}$
\end{algorithmic}
\end{algorithm}

\begin{algorithm}
\caption{Calculation of Backward Transfer}\label{alg:bt}
\begin{algorithmic}
\Require Task set $\mathcal{T}$
\Require Evaluation Block Performances $\{P_E(n,t)\}$ for $n=1,...,N$, $t\in\mathcal{T}$
\Ensure $BackwardTransfer$
\vspace{0.2cm}
\State $BTs = LearnedTasks = LearnedTaskPairs = \emptyset$
\For{Learning Blocks $n=2,...,N$}
\If{$t(n) \not\in LearnedTasks$}
\State $LearnedTasks \gets LearnedTasks \cup \{t(n)\}$
\EndIf
\For{Tasks $t \in \mathcal{T}\setminus\{t\}$}
\If{$\{t(n), t\}\not\in LearnedTaskPairs$ and $t \in LearnedTasks$}
\State $LearnedTaskPairs \gets LearnedTaskPairs \cup \{\{t(n), t\}\}$
\State $P_{n-1}, P_n = P_E(n-1, t), P_E(n, t)$
\State $BTs \gets BTs\cup\{\mathrm{Contrast}(P_n, P_{n-1})\}$
\EndIf
\EndFor
\EndFor
\State $BackwardTransfer \gets \mathrm{mean}\{BTs\}$
\end{algorithmic}
\end{algorithm}

\begin{algorithm}
\caption{Calculation of Performance Relative to a Single Task Expert}\label{alg:rp}
\begin{algorithmic}
\Require Task set $\mathcal{T}$
\Require Learning Block Performances $\{P_L(n,t,\ell)\}$ for $\ell=1,...,\ell(n)$, $n=0,...,N$, $t\in\mathcal{T}$
\Require STE Performances $\{P_{STE}(n,t,\ell)\}$ for $\ell=1,...,\ell(n)$, $n=0,...,N$, $t\in\mathcal{T}$
\Ensure $RelativePerformance$
\vspace{0.2cm}
\State $RPs = \emptyset$\Comment{Relative performances for each task}
\For{Tasks $t \in \mathcal{T}$}
\State $RP_t \gets \dfrac{\sum_{n=1}^N\sum_{\ell=1}^{\ell(n)} P_L(n,t,\ell)}{\sum_{n=1}^N\sum_{\ell=1}^{\ell(n)} P_{STE}(n,t,\ell)}$
\vspace{0.2cm}
\State $RPs \gets RPs \cup \{RP_t\}$
\EndFor
\State $RelativePerformance \gets \mathrm{mean}\{RPs\}$
\end{algorithmic}
\end{algorithm}

\begin{algorithm}
\caption{Calculation of Sample Efficiency}\label{alg:se}
\begin{algorithmic}
\Require Task set $\mathcal{T}$
\Require Learning Block Performances $\{P_L(n,t,\ell)\}$ for $\ell=1,...,\ell(n)$, $n=0,...,N$, $t\in\mathcal{T}$
\Require STE Performances $\{P_{STE}(n,t,\ell)\}$ for $\ell=1,...,\ell(n)$, $n=0,...,N$, $t\in\mathcal{T}$
\Require Smoothing function $\mathrm{Smooth}$, Window length $w$
\Ensure $SampleEfficiency$
\vspace{0.2cm}
\State $SEs = \emptyset$\Comment{Sample efficiency scores for each task}
\For{Task $t \in \mathcal{T}$}
\State\Comment{Concatenate all learning blocks for the current task $t$}
\vspace{0.2cm}
\State $P_{L,\text{cat},t} = \mathrm{concat}(P_L(n,t) : t(n) = t)$
\vspace{0.2cm}
\State $P_{STE,\text{cat},t} = \mathrm{concat}(P_{STE}(n,t) : t(n) = t)$
\vspace{0.2cm}
\State $\tilde{P}_{L,\text{cat},t},\tilde{P}_{STE,\text{cat},t} = \mathrm{Smooth}(P_{L,\text{cat},t}),\mathrm{Smooth}(P_{STE,\text{cat},t})$
\\\Comment{Find saturation performance values and experience locations}
\State $SatVal(L,t), SatExp(L,t) = \max \tilde{P}_{L,\text{cat},t},\arg\max \tilde{P}_{L,\text{cat},t}$
\vspace{0.2cm}
\State $SatVal(STE,t), SatExp(STE,t) = \max \tilde{P}_{STE,\text{cat},t},\arg\max \tilde{P}_{STE,\text{cat},t}$
\vspace{0.2cm}
\State $SEs \gets SEs \cup \left\{\dfrac{SatVal(L, t)}{SatVal(STE,t)}\dfrac{SatExp(STE,t)}{SatExp(P,t)}\right\}$
\EndFor
\State $SampleEfficiency \gets \mathrm{mean}\{SEs\}$
\end{algorithmic}
\end{algorithm}

\begin{algorithm}
\caption{Calculation of Performance Maintenance}\label{alg:pm}
\begin{algorithmic}
\Require Task set $\mathcal{T}$
\Require Evaluation Block Performances $\{P_E(n,t,\ell)\}$ for $\ell=1,...,\ell(n)$, $n=0,...,N$, $t\in\mathcal{T}$
\Ensure $PerformanceMaintenance$
\vspace{0.2cm}
\State $MVs(t) = \emptyset$ for all $t \in \mathcal{T}$\Comment{Maintenance Values}
\State $PMs = \emptyset$\Comment{Performance Maintenance scores}
\State $MRB(t) = -\infty$ for all $t \in \mathcal{T}$\Comment{Most recent LB index for each task}
\For{Learning Block $n=1,\hdots,N$}
\State $MRB(t(n)) = n$
\For{Task $t \in \mathcal{T}$}
\If{$MRB(t) > 0$ and $t \neq t(n)$}
\State $MVs(t) \gets MVs(t) \cup \{P_E(n, t) - P_E(MRB(t), t)\}$
\EndIf
\EndFor
\EndFor
\For{Task $t \in \mathcal{T}$}
\State $PMs \gets PMs \cup \{\mathrm{mean}\{MV(t)\}\}$
\EndFor
\State $PerformanceMaintenance \gets \mathrm{mean}\{PMs\}$
\end{algorithmic}
\end{algorithm}

%%%%%%%%%%%%%%%%%%%

\section{Statistical Reliability}\label{sec:reliability}

Statistical analyses can fail to recognize when two algorithms evaluated on the same benchmark are the same algorithm \citep{Colas2018Power}. Varying approaches have been recommended to mitigate this, including the use of the almost stochastic dominance test \citep{Dror2019Dominance} and performance profiles during training \citep{Agarwal2021deep}.

In Figure~\ref{fig:scenario-design}, we present a nominal \gls{LL} scenario. An agent is sent through a sequence of tasks; at the end of each lifetime, it generates a set of LL metrics. This design suggests two questions: (1) How should $K$ (the number of repetitions) be chosen ahead of time? and (2) How should metrics be aggregated across lifetimes after the fact?

Reliably assessing the variability in the responses of the agent, assuming the inherent variability of its inputs, requires assessing performance of the agent over multiple lifetimes. We outline a procedure based on guidance in \cite{NISTMethods} and similar to \cite{Colas2018Power} to determine the number of lifetimes that need to be run, for a given Evaluation Protocol, to assess an agent's performance.

For a given evaluation protocol, let $Y$ be the random variable of values a metric can take, assumed to follow a normal distribution with population mean and standard deviation $\mu$ and $\sigma$. We seek to characterize a system's performance by estimating $\mu$. For an estimator $\bar{Y}$ of $Y$ (typically, the sample mean of a set of values of the metric taken from multiple independent runs), we evaluate the null hypothesis that the error in estimating $|\bar{Y} - \mu|$ is no more than some error threshold $\delta$. Our hypothesis of normality is strong, but it is meant to enable easy and efficient estimation of distribution properties, as well as assumptions that can be checked in practice.

One option is to choose a threshold $\delta$ based on the specific Protocol. However, the space of potential protocols is vast, even for a relatively small number of scenario tasks and agent configurations, and there is no guarantee that the same threshold will be informative across protocols. We follow common practice and choose the error threshold to be defined as a multiple of the standard deviation: $\delta = k \sigma$. Thus, a procedure for determining required sample size prior to training any agents is specified by the choice of the multiple $k$, the type I error rate $\alpha$, and the type II error rate $\beta$. We recommend, as a default, setting $k = 1, \alpha = 0.05$, and $\beta = 0.1$. This yields a suggested required sample size of at least 11 runs. Evidence from works such as \cite{Agarwal2021deep} suggests this is likely an underestimate and so, if computational resources and time allow, more data will be of value.

With respect to the second question, we recommend two procedures for comparing the distribution of an agent's metric values to some threshold. The student $t$-test can be used to compare raw distributions of metrics values. However, this approach can be unreliable in the case that the values of a metric are highly non-normal (from, e.g., outliers or skewness). In that case, a more robust alternative is to binarize values by checking if they surpass that threshold and performing a statistical test on that set of binary values.

\section{Summary of Tasks used in SG Case Studies}

\begin{table}[H]
\centering
\resizebox{\textwidth}{!}{
\begin{tabular}{|c|c|c|}
\hline
System   Group & Environment   & Task Descriptions                                             \\ \hline
UPenn (\cref{sec:sg-upenn})          & AI Habitat   & Classify/Find Seating Furniture                   \\
               &              & Classify/Find Plumbing Furniture                  \\
               &              & Classify/Find Large Furniture                     \\ \hline
Teledyne (\cref{sec:sg-teledyne})      & AirSim       & Classify Emergency Management Assets, low altitude       \\
               &    Drone     & Classify Emergency Management Assets, high altitude    \\
               &              & Classify Dept. of Transportation Assets, low altitude \\ \hline
HRL (\cref{sec:sg-hrl})           & CARLA        & Car navigation                                    \\
               &              & Motorcycle navigation                             \\
               &              & Motorcycle navigation, opposite lane              \\ \hline
ANL (\cref{sec:sg-argonne})           & L2Explorer   & Identify targets                                  \\
               &              & Navigation despite distractors                    \\
               &              & Forage specific resources                         \\ \hline
SRI (\cref{sec:sg-sri})           & StarCraft II & Collect Resources                                 \\
               &              & Defeat Large Enemies                              \\
               &              & Defeat Small Enemies  \\ \hline                           
\end{tabular}
}
\caption{High level task descriptions used in the five case studies discussed in \cref{sec:case_studies}. Note that since the UPenn group performed both classification and \gls{RL} experiments, their tasks involved either classifying or finding, respectively. }
\label{tab:envs-and-tasks}
\end{table}

\section{T-Test Values for SG Case Study Data}\label{sec:t-test-vals}

\begin{table}
\scalebox{0.85} {
\begin{tabular}{|p{0.15\linewidth}|p{0.23\linewidth}|p{0.15\linewidth}|p{0.15\linewidth}|p{0.15\linewidth}|p{0.15\linewidth}|p{0.15\linewidth}|}
    \hline
        SG & Config & PM & FT & BT & RP  & SE \\ \hline
        Argonne & L2Explorer & -1.31 & 8.65 & 2.93 & 1.20 & 1.20 \\ 
        & Roundworld & 5.20 & 6.93 & 2.80 & 4.08 & 16.63 \\ \hline
        HRL & Condensed & -0.23 & 12.67 & 4.00 & 6.42 & 3.67 \\ 
        & Disp & -3.77& 18.78 & 4.10 & 6.49 & 9.06 \\
        & SCP Ablation & 0.25 & 10.75 & 3.45 & 9.10 & 5.87 \\ \hline
        SRI & M12 Condensed & -5.13 & 8.31 & -0.12 & -2.36 & 6.77 \\
        & M15 Condensed & -5.52 & 8.31 & 2.46 & 4.57 & 15.31 \\
        & M18 Condensed & -6.73 & 15.31 & 0.62 & 5.72 & 10.95 \\
        & M12 Alternating & -4.66 & 1.05 & -3.53 & -5.57 & 6.15 \\
        & M15 Alternating & -5.44 & 1.15 & -5.09 & -4.84 & 8.51 \\
        & M18 Alternating & -2.91 & 2.13 & -2.31 & -2.52 & 5.09 \\ \hline
        Teledyne & C5 Ablated & 1.98 & 79.33 & 0.72 & 2.71 & 3.55 \\
        & UML & 2.82 & 72.01 & 3.15 & 1.80 & 3.13 \\ \hline
        UPenn & DF-CNN & -2.36 & 0.63 & -2.24 & 22.00 & 29.41 \\
        & META-KFO & -8.20 & NaN & -7.99 & 20.83 & 31.47 \\
        & RL M12 & -5.59 & -0.28 & 0.26 & -7.14 & -2.52 \\
        & RL M15 & -2.37 & 3.39 & 4.11 & -14.43 & 1.56 \\
        & RL M18 & 0.99 & 2.97 & 5.21 & -13.27 & -18.79 \\ \hline
\end{tabular}
} % End scale box
    \caption{t values from a one-tailed t-test to determine whether the value is significantly greater than the LL Threshold value for that metric.
    Note that the UPenn META-KFO system is designed to speed up the rate of adapting to a new task, but this does not happen until data for that task is seen, leading to unchanged task values and a standard deviation of zero for a jumpstart formulation of FT.}
\label{tab:metric-t-tests-t-values}
\end{table}

\section{Computational Costs of Lifelong Learning}\label{sec:cc}

Different \gls{LL} algorithms can potentially have different computational costs. For example, an algorithm with experience replay might be more computationally expensive during deployment than one that grew the network as needed. Unfortunately, it is challenging to compare these costs across agents given differences in learning frameworks, distributed training, and environments. Instead, we attempted to get insight into $CostOverhead$, the relative cost imposed by an \gls{LL} system as it tries to preserve and transfer learning across multiple tasks, compared to the same algorithm being applied to just a single task (see Table \ref{tab:cost-overhead-definition}). For instance, $CostOverhead=1.5$ indicates that it takes 1.5x more computational effort to process a single learning experience (LX) during deployment (when learning multiple tasks) compared to a single-task setting.

\begin{table}
\scalebox{0.85} {
\begin{tabular}{|p{8.5cm}|p{6.5cm}|}
\hline
$ RawCost^{multitask}$ & Elapsed time for a single lifetime with multiple tasks, averaged across the submitted runs.\\ 
\hline
$ RawCost^{singletask}$ & Elapsed time for the single task expert, trained to saturation. \\ 
\hline
$ CostPerLX^{multitask} = \frac{RawCost^{multitask}}{Total\:Number\:of\:LXs^{multitask}} $ & Time Cost per LX, for the multi-task lifelong learner \\
\hline
$ CostPerLX^{singletask} = \frac{RawCost^{singletask}}{Total\:Number\:of\:LXs^{singletask}} $ & Time Cost per LX for the single-task expert\\
\hline
$ CostOverhead = \frac{CostPerLX^{multitask}}{CostPerLX^{singletask}} $ & Cost overhead of lifelong learning \\
\hline
\end{tabular}
} % End scale box
\caption{Definition of $CostOverhead$. Note that $RawCost$ and $CostPerLX$ (both single and multitask) are measured in seconds}
\label{tab:cost-overhead-definition}
\end{table}

It should be noted that $CostOverhead$ is a crude measure, with several limitations: it does not distinguish between learning and evaluation experiences, does not take overall performance into account, and does not separately consider agent and environment computation (for example, a complex 3D environment like AirSim may take more computational resources to render than StarCraft). Even so, $CostOverhead$ can provide useful insight. When applied to preliminary versions of the LL algorithms developed by the SGs, the $CostOverhead$s ranged from 1.27 to 2.53, indicating that some LL algorithms potentially had twice the multi-task overhead of others. Notably, the $CostOverhead$s are contained within a small band of values, which is remarkable given the diversity of environments, tasks and learning algorithms.

\newpage
\bibliographystyle{elsarticle-harv} 

\bibliography{bib_teledyne,bib_upenn,bib_metrics,bib_vandy,bib_sri,bib_anl}

\end{document}